\documentclass[10pt, a4paper]{article}

\usepackage{lrec2026} 
\usepackage{xcolor}
\usepackage{comment}
\usepackage{url}

\usepackage[T1]{fontenc}
\usepackage[utf8]{inputenc}
\usepackage{listings}
\usepackage{subcaption} 

\usepackage[hang,flushmargin]{footmisc} 

\makeatletter
\renewcommand\footnotesize{\@setfontsize\footnotesize{9pt}{10pt}}
\makeatother

\title{SPOT: An Annotated French Corpus and Benchmark for Detecting Critical Interventions in Online Conversations}

\name{Manon Berriche\textsuperscript{*,1}, Célia Nouri\textsuperscript{*,1,2}, Chloé Clavel\textsuperscript{2,3}, Jean-Philippe Cointet\textsuperscript{1}} 

\address{\textsuperscript{*}Equal contributions, \textsuperscript{1}Sciences Po, médialab, \textsuperscript{2}INRIA, ALMAnaCH,  \textsuperscript{3}Télécom Paris \\
\texttt{\{manon.berriche, celia.nouri\}@sciencespo.fr}}

\abstract{We introduce \textsc{SPOT} (\textit{Stopping Points in Online Threads}), the first annotated corpus translating the sociological concept of \textit{stopping point} into a reproducible NLP task. Stopping points are ordinary critical interventions that pause or redirect online discussions through a range of forms — irony, subtle doubt or fragmentary arguments— that frameworks like counterspeech or social correction often overlook. We operationalize this concept as a binary classification task and provide reliable annotation guidelines. The corpus contains 43{,}305 manually annotated French Facebook comments linked to URLs flagged as false information by social media users, enriched with contextual metadata (article, post, parent comment, page or group, and source). We benchmark fine-tuned encoder models (\textsc{CamemBERT}) and instruction-tuned LLMs under various prompting strategies. Results show that fine-tuned encoders outperform prompted LLMs in $F_1$ score by more than 10 percentage points, confirming the importance of supervised learning for emerging non-English social media tasks. Incorporating contextual metadata further improves encoder models $F_1$ scores from $0.75$ to $0.78$. We release the anonymized dataset, along with the annotation guidelines and code in our \href{https://github.com/celia-nouri/SPOT-benchmark}{code repository}, to foster transparency and reproducible research.
 \\ \newline  }

\begin{document}

\maketitleabstract

\section{Introduction}

Research on online discourse has largely focused on phenomena perceived as harmful — such as polarization, misinformation or hate speech — and on their automated detection and measurement \citep{delvicario2016spreading,Waseem_2016, shu2017fake, vosoughi2018spread}. A smaller but growing literature has examined how users respond to these harms through counter-speech, social correction, or user-led moderation \citep{buerger2021,Falk_Vecchi_Jundi_Lapesa_2024,bode2024user}. In parallel, the Natural Language Processing (NLP) community has produced annotated datasets and benchmarks addressing related phenomena—from stance and (dis)agreement to counter-speech and corrective replies—along with annotation schemes and models designed to capture such behaviors \citep{kuccuk2020stance, Bonaldi_Chung_Abercrombie_Guerini_2024}. 

Yet much of this work tends to focus on explicit, goal-directed interventions (evidence-based refutations or collective moderation) while overlooking more common everyday reactions that do not fully correct or sanction a message but nevertheless interrupt, reframe, or stall its circulation. Still, they matter as they reveal forms of criticism that standard taxonomies neglect and provide empirical windows into how communities self-negotiate norms and interpretive frames within a thread. However, they pose challenges for NLP as these interventions are often subtle, ironic, or fragmentary.

\begin{figure}[!ht]
\centering
\begin{subfigure}[t]{\columnwidth}
    \centering
    \includegraphics[width=\linewidth]{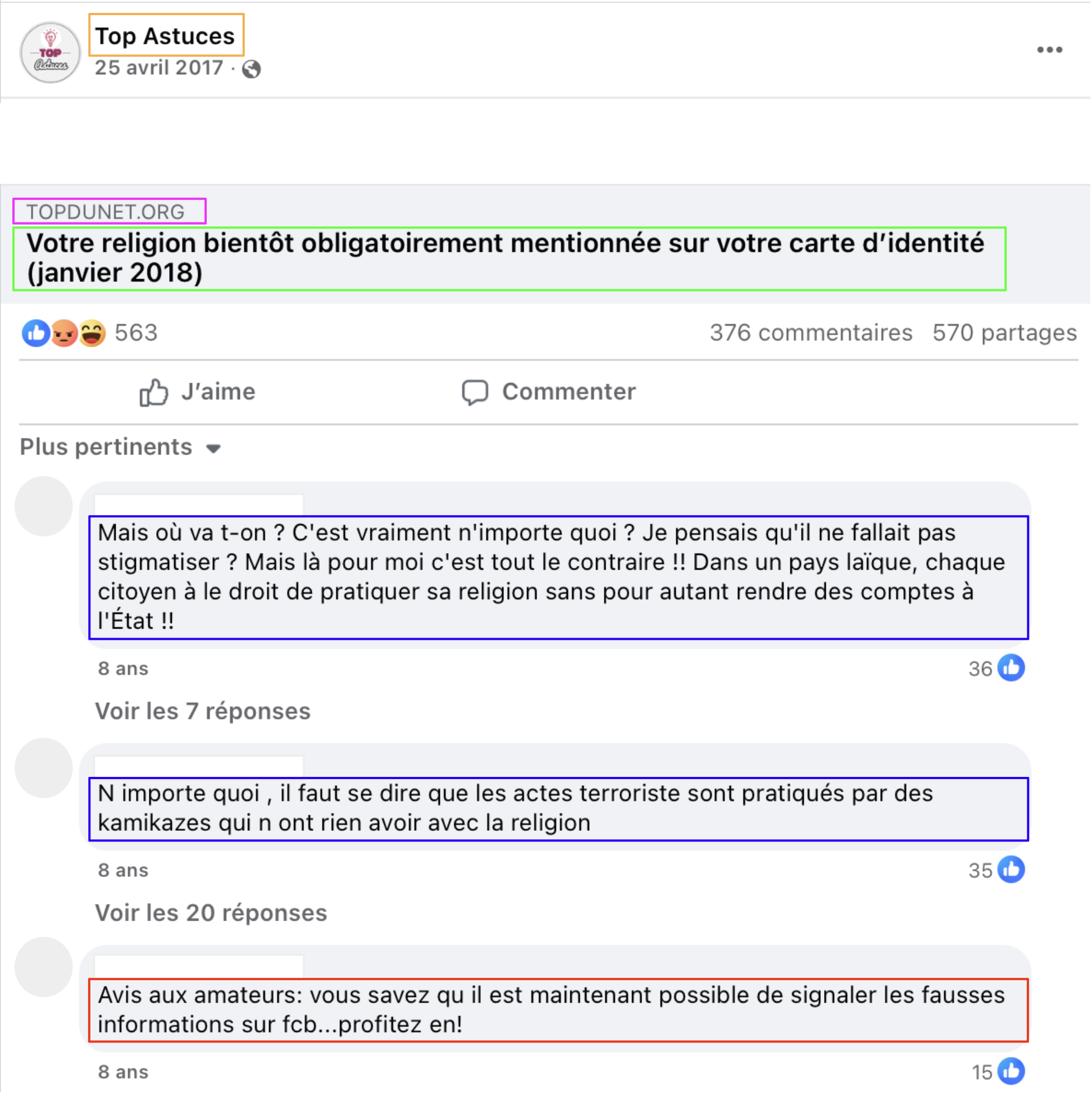}
    \caption{Top-level post and comments.}
    \label{fig:stop}
\end{subfigure}

\vspace{0.3em} 

\begin{subfigure}[t]{\columnwidth}
    \centering
    \includegraphics[width=\linewidth]{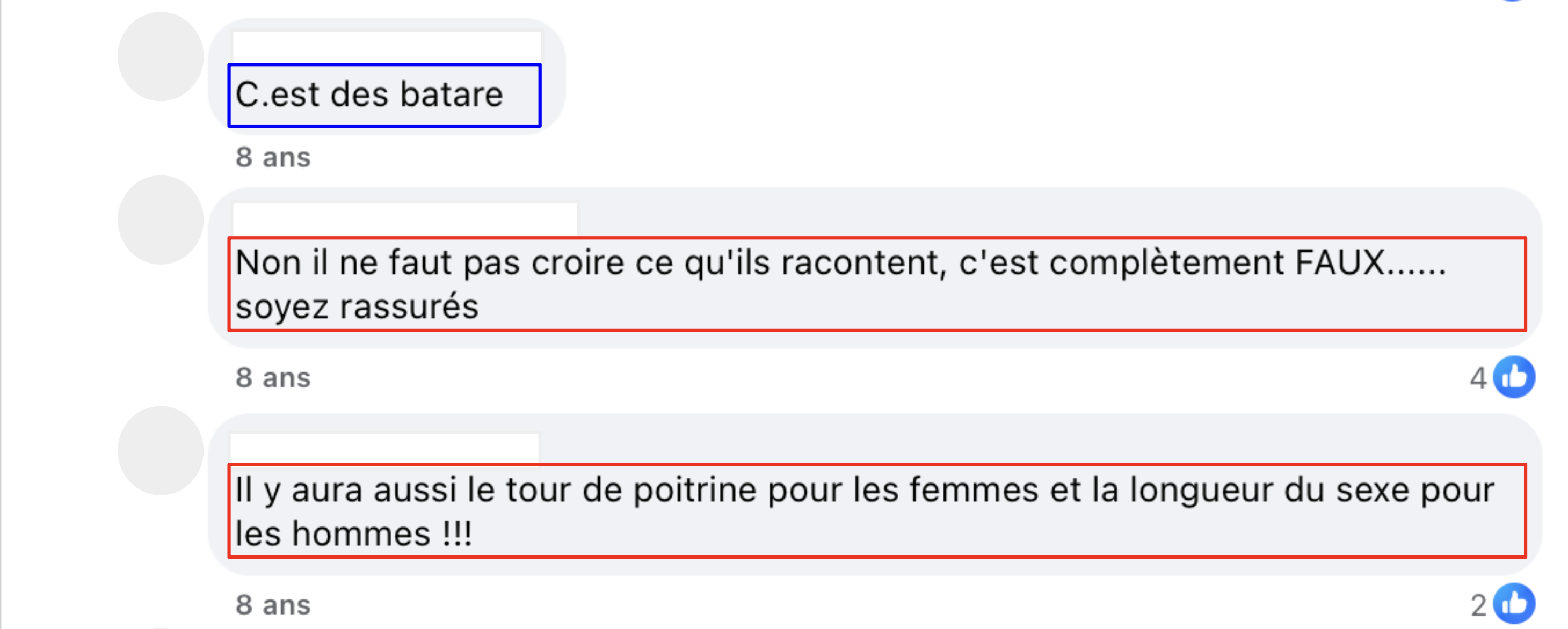}
    \caption{Replies under the first top-level comment.}
    \label{fig:replies}
\end{subfigure}
\caption{An example thread from the SPOT corpus showing  (a) the post with top-level comments, and (b) the reply thread under the first comment. 
Colors indicate page/group name (orange), domain name (pink), article title (green), non-stopping points (blue), and stopping points (red). 
English translations for (a)\protect\footnotemark[1] and (b)\protect\footnotemark[2] are provided below.}
\label{fig:spot-example}
\end{figure}

To address this gap, we introduce and operationalize the sociological notion of a \emph{stopping point} \citep{berriche:tel-05409923}: an ordinary critical intervention that marks hesitation, resistance, or creates a pause or shift in an online conversation without necessarily resolving the factual status of the contested content. Examples range from skeptical prompts (\textit{“Is this true?”}) and dismissive asides (\textit{“You’re talking nonsense”}) to terse denunciations (\textit{“Report”}) or corrective replies with links, and can also take ironic forms (\textit{“When pigs fly”}, \textit{“Yeah, and I’m the Queen of England”}). Crucially, stopping points are defined by their \textit{conversational function}—momentarily halting or redirecting the flow of interaction—rather than by tone, polarity, or factual accuracy, making them particularly challenging for both annotation and automatic detection. Indeed, as illustrated by Figure \ref{fig:stop}, lexical cues alone are not sufficient to identify stopping points. Although the first two comments (\textit{“This is completely absurd!”}, \textit{“This is ridiculous”}) appear critical in tone, they take the post at face value. They therefore express agreement rather than challenge or redirection and should be annotated as non–stopping points. In contrast, the third comment (\textit{“Attention everyone: did you know that it’s now possible to report false information on Facebook?”}) reframes the discussion as a moderation issue and invites collective action, thus qualifying as a stopping point.
In Figure~\ref{fig:replies}, the third reply (\textit{“There will also be women’s bust sizes and men’s penis lengths!!”}), replying to the first comment in Figure~\ref{fig:stop}, differs from a straightforward rebuttal but still serves as a stopping point. Without explicit critical markers, it uses irony and hyperbole to recast the original claim as an absurd slippery-slope caricature. This performative reframing contrasts with goal-directed interventions like social correction or counterspeech, which typically provide evidence or reasoned arguments to counter misinformation or hate speech. 
Together, these examples underscore that reliably annotating and detecting stopping points requires contextual understanding of the broader discussion, rather than sole reliance on lexical cues from isolated comments.

\footnotetext[1]{\textbf{Page/Group name}: Top Tips; \textbf{Domain name}: TopsFromTheWeb.org; \textbf{Article title}: Your religion will soon have to be stated on your ID card (January 2018); \textbf{Comment 1}: But where are we headed? This is completely absurd! I thought we weren't supposed to stigmatize people? But to me, here, it's the exact opposite!! In a secular country, every citizen has the right to practice their religion without having to answer to the government!!; \textbf{Comment 2}: This is ridiculous, terrorist acts are carried out by suicide bombers who have nothing to do with religion; \textbf{Comment 3}: Attention everyone: did you know that it’s now possible to report false information on Facebook? Take advantage of it!}

\footnotetext[2]{\textbf{Reply 1} Fuckers; \textbf{Reply 2} No, don't believe what they're saying, it's completely FALSE... rest assured; \textbf{Reply 3} There will also be women's bust sizes and men's penis lengths!!!}
This paper makes four contributions. First, we translate the sociological notion of stopping point into a reproducible annotation task and provide detailed annotation guidelines. Second, we present SPOT, a corpus of 43{,}305 manually annotated French Facebook comments linked to URLs flagged as false information. The dataset is enriched with parent-post and comment context, shared URL and source information, page or group features. Third, we benchmark automated approaches for the stopping point detection task by comparing a fine-tuned \textsc{CamemBERT} model \citep{Martin_Muller_Suárez_Dupont_Romary_Clergerie_Seddah_Sagot_2020}, trained with different combinations of textual and contextual features, to instruction-tuned Large Language Models (LLMs) evaluated through various prompting strategies. Our results show that fine-tuned encoders substantially outperform prompted LLMs on the SPOT corpus, underscoring the effectiveness of supervised domain adaptation for social media text analysis in non-English settings. We also find that incorporating the publication context significantly improves performance. Fourth, we analyze the types of case that are incorrectly predicted by current methods, providing empirical guidance for future work on improving modeling strategies for social media text classification tasks.

Taken together, these resources and results suggest that modeling everyday critical interventions in online conversations requires moving beyond lexical cues to integrate conversational and social media context, including information about the post, the source, and the hosting page or group.

\section{Related Work and Conceptual Framework}

\subsection{Studies on User Critical Interventions}

\paragraph{Related concepts.} A growing body of research examines how ordinary users contest, correct, or contain problematic content through incremental, everyday critical interventions rather than through formal fact-checking or take-down procedures. These practices have been variously labeled as social correction \citep{bode2024user}, counter-speech \citep{buerger2021}, online civic interventions \citep{porten2020online}, informal social control \citep{watson2019will}, or user moderation \citep{Falk_Vecchi_Jundi_Lapesa_2024}. These critical interventions are broadly understood as reactive to media or user content perceived as harmful or problematic and seek to mitigate its negative effects on the public sphere \citep[p.~733]{ziegele2020lonely}. 

\paragraph{Related datasets.} Researchers have also assembled a range of annotated resources addressing related phenomena. Early dialogue and forum corpora (LiveJournal, Wikipedia Talk Pages, IAC) documented agreement/disagreement \citep{andreas2012annotating,bender2011annotating,walker2012corpus}, while social-media collections (Coarse Discourse, DEBAGREEMENT) extended this work to Reddit \citep{zhang2017characterizing,pougue2021debagreement}. Counterspeech corpora (CONAN family, MultiTarget-CONAN, DIALOCONAN) provide expert-crafted or synthetic counter-narratives to hateful content \citep{Chung_Kuzmenko_Tekiroglu_Guerini_2019, Fanton_Bonaldi_Tekiroglu_Guerini_2021, Bonaldi_Chung_Abercrombie_Guerini_2024, poudhar2024strategy}. Datasets oriented toward misinformation correction and rumor resolution include PHEME, RumourEval, Emergent, Twitter15/16 and several COVID-19 correction corpora \citep{zubiaga2016pheme,derczynski2017semeval,gorrell2018rumoureval,ferreira2016emergent,ma2017detect,ma2023characterizing}. Finally, everyday user moderation has been examined in smaller datasets such as the $1{,}000$ comment–reply pairs of UMOD \citep{Falk_Vecchi_Jundi_Lapesa_2024}.

\paragraph{Limits.} Two recurring limitations constrain the use of existing work for studying ordinary critical interventions online. First, many studies rely on narrow and normatively loaded definitions of user interventions—considering as “critical” only those that resemble expert moderation or display clear disagreement, argumentation, or evidence. This framing overlooks the diversity of ordinary user reactions—irony, doubt, resistance, or indifference—that also shape how problematic content circulation is hampered. It further assumes a normative model of deliberation, evaluating interventions by their “success” in correcting misinformation or improving discussion quality, and thereby presuming that they are necessarily correct, rational, or appropriate. Such assumptions overlook a key pragmatic fact: critical interventions may be mistaken, poorly timed, or perceived as hostile. The second limitation is that most existing datasets are English-centric and focus on isolated units (sentences or turns), ignoring thread structure and contextual dependencies.

\subsection{Conceptual Foundations and Definition of Stopping Point}
\label{subsec:def}

To address the conceptual limitations of prior frameworks, the notion of \emph{stopping point} was introduced by \citet{berriche:tel-05409923} as a sociologically grounded approach to studying everyday critical interventions in online discussions. It builds on three complementary traditions—reception studies, pragmatic sociology, and conversation analysis—that conceive meaning and disagreement as situated, interactional processes rather than stable expressions of opinion. Reception studies emphasize that audiences actively interpret and sometimes contest media messages according to their knowledge and context \citep{Hall_1980, Lull_1995}. Pragmatic sociology highlights that users mobilize diverse forms of critique—moral, epistemic, and procedural—when expressing disagreement \citep{Boltanski_Thevenot_2006, Boltanski_2011}. Conversation analysis shows that a comment’s pragmatic force depends on its sequential position in a thread  \citep{Sacks_1974, Heritage_1984, Hutchby_Wooffitt_2008}. Taken together, these perspectives frame critical interventions as context-dependent acts that derive meaning from their relation to surrounding turns rather than from textual content alone.

Drawing on this framework, a stopping point is defined as a user’s critical intervention in an online discussion. It can cover a range of forms, from a brief expression of doubt or a dismissive remark to ironic responses or more elaborate refutation with counter-arguments or sources. 

The act itself does not imply the factual accuracy or normative legitimacy of the intervention. A stopping point thus refers to the \emph{function} of an utterance within a conversation rather than its accuracy or rhetorical form. Tone, style, or hostility do not disqualify an intervention: sarcastic or emotional comments can still act as stopping points if they indicate hesitation, resistance, or refusal. These interventions, even when non-constructive, momentarily halt the circulation of content and influence the negotiation of meaning and norms within the thread. Crucially, a stopping point is a \emph{speech act} rather than a stable stance: the same user may alternate between aligned and critical turns, retract or reinforce previous comments. Stopping points are therefore best understood as conversational pauses—brief interruptions that can trigger rebuttals, clarifications, or further escalation.

Methodologically, this perspective implies that identifying stopping points requires thread-level context (media source, page, article, parent comment) rather than treating comments in isolation.

\section{SPOT Corpus}

\subsection{Data Collection}

The SPOT corpus was constructed using data derived from the Facebook Privacy Protected Shared URLs Dataset, which is accessible to researchers through the Social Science One consortium \citep{Messing_DeGregorio_Hillenbrand_King_Mahanti_Mukerjee_Nayak_Persily_State_Wilkins_2023}. The Shared URLs Dataset contains around 38 million URLs that were publicly shared at least 100 times on Facebook between January 2017 and July 2019. Each URL is associated with a certain number of metrics allowing to measure how their audience engaged with them. Among those metrics, the dataset allows to measure how many times a URL was signaled as “fake”.

For the construction of SPOT, we selected only URLs reported as “fake” and shared on public French Facebook pages or groups, which resulted in a subset of 904 flagged URLs. Importantly, these URLs were user-flagged, meaning that they represent content perceived as potentially false rather than verified as false by professional fact-checkers. They should therefore be understood as claims of uncertain epistemic status. This sampling strategy offers a privileged vantage point from which we can capture naturally occurring discussions in which stopping points emerge in response to exposure to misleading or contentious content.

All corresponding Facebook posts sharing these links—representing a total of 30,157 posts—and their associated discussion threads were collected using the \textit{Minet} web-mining tool \citep{plique2019minet}. In total, the dataset includes 441{,}149 comments authored by 294{,}988 unique users. 

\subsection{Exploratory Observations and Annotation Guidelines}

Automating the detection of stopping points required manual annotations, themselves grounded in a robust operational definition and explicit decision rules. To develop these guidelines, one of the authors (trained in sociology and qualitative methods) conducted an immersive online ethnography across a first limited sample of Facebook pages and groups (\(n \approx 50\)) that shared user-flagged URLs. This fieldwork consisted in the systematic reading of full threads, the comparative notes in a field journal, and the identification of recurring interactional formats and borderline cases. These observations served three key functions.

First, they made the stopping point concept operational by (i) fixing the unit of analysis (the individual comment or nested reply) and (ii) enumerating the typical targets of criticism, such as content credibility (\textit{“This is false”}), source reliability (\textit{“this website is biased”}), form/media (\textit{“photoshopped image”}), the poster (\textit{“I’m unfollowing”}), or other users (\textit{“can’t believe people fall for this”}). Second, they dictated a context-first annotation procedure requiring annotators to (i) open the shared URL and read the parent post, (ii) examine at least the immediately adjacent turns (one above, one below when available), and (iii) consult the page/group description when local norms or community identity seemed relevant. Third, the field notes surfaced difficult cases (implicit refutation, ironic endorsement, link-only replies, and ultra-short fragments such as single words or emojis), which were documented and used to craft concrete resolution rules and illustrative examples.

The resulting annotation guide encodes: (i) a compact operational definition of a stopping point and (ii) explicit decision rules for ambiguous cases (irony, nested replies, link-only responses). The full guidelines, including annotated examples drawn from the field journal, are available in the project repository.

\subsection{Guidelines Validation, Annotation and Inter-Rater Reliability}
\label{sec:annotation}

As part of an iterative calibration, the draft guidelines were tested for comprehensibility by three annotators, who independently coded ten challenging cases. Discrepancies were reconciled through discussion, and the decision rules were updated accordingly (see Appendix~\ref{appendix:annotation} for details), resulting in the finalized guidelines. These guidelines were then applied to annotate the entire dataset.

The main annotation sample contains 43{,}305 comments (10\% of the collected Facebook corpus), drawn from 1{,}061 randomly selected posts. For each post, all associated comments were annotated within the full conversation to preserve thread context. One of the lead authors (trained in sociology and qualitative methods) completed the initial annotation of the full sample following the calibrated guidelines.

Finally, to assess the reproducibility of these annotations, a validation subset of 500 comments was independently annotated by two additional trained experts with backgrounds in Sociology and Natural Language Processing. These comments were selected from a random sample of posts, with a maximum of five randomly chosen comments per post to ensure diversity of situations and topics. Inter-rater reliability (IRR) was quantified to ensure annotation quality. Since our task involves binary categorical data fully annotated by three raters, we report both \textit{Fleiss' $\kappa$} \cite{fleiss1971measuring} and \textit{Krippendorff's $\alpha$} \cite{krippendorff2004content}. Fleiss' $\kappa$ provides a standard measure of chance-corrected agreement for fully annotated categorical data, while Krippendorff's $\alpha$ is broadly recommended and reported for small, expert-coded datasets in computational social science and NLP. We also report raw percent agreement for interpretability. 

The obtained reliability coefficient of $\alpha \approx 0.80$ indicates a robust and substantial agreement for a binary annotation task. According to \citet{krippendorff2004content}, values of $\alpha \geq 0.8$ indicate strong reliability. Similarly, Fleiss’ $\kappa$ values above $0.61$ indicate substantial agreement, while those above $0.81$ indicate almost perfect agreement \citep{landis1977measurement}. Overall, these strong coefficients  affirm that the annotation process achieved a high degree of consistency, ensuring the creation of a reliable and high-quality gold standard dataset.
To provide robust estimates of variability, we computed 95\% confidence intervals for all IRR metrics using bootstrap resampling: 500 bootstrap samples were drawn with replacement from the subset annotated by the three annotators, and the IRR metrics were recalculated for each sample. The resulting intervals are reported alongside point estimates in Table~\ref{tab:agreement}.The labels from the first annotator were used as the final gold standard for subsequent model training and evaluation.

\begin{table}[ht!]
\centering
\small
\setlength{\tabcolsep}{4pt}
\begin{tabular}{lcc}
\hline
\textbf{Metric} & \textbf{Value} & \textbf{95\% CI} \\
\hline
Raw agreement & 0.9067 & [0.8829, 0.9306] \\
Fleiss’ $\kappa$ & 0.8036 & [0.7494, 0.8578] \\
Krippendorff’s $\alpha$ & 0.8037 & [0.7496, 0.8579] \\
\hline
\end{tabular}
\caption{Inter-annotator agreement on the 500-sample dataset, with bootstrap 95\% confidence intervals.}
\label{tab:agreement}
\end{table}

\subsection{Corpus Description and Availability}

The SPOT Corpus contains 43{,}305 manually annotated comments drawn from 1{,}061 posts and 253 shared URLs, published across 275 public French Facebook pages and groups. Each comment is linked to its post, the shared article (URL, title, description), and the hosting community (page or group name). 

Each post includes several elements forming its \textit{publication context} (Figure~\ref{fig:stop}): the \textit{account name}, which provides cues about the page or group’s thematic or ideological orientation), the \textit{post message} (which may reproduce, criticize, or comment on the article title), and metadata about the shared article (domain name, title, description). Comments can be direct comments to the post (28{,}457; 65.7\%) or replies to another comment (14{,}848; 34.3\%), in which case the parent comment is part of the publication context (Figure~\ref{fig:replies})

Overall, 4{,}306 comments (9.9\%) were labeled as stopping points. SPOT provides a large-scale, conversation-level resource for studying how users collectively problematize, contest, or nuance potentially misleading content in authentic social media contexts.

The SPOT corpus is made available to the research community upon request through a secure institutional data repository. Access is granted only for academic purposes after evaluation of the research project, ensuring both reproducibility and the protection of users’ privacy.

\section{Classification Task}

\subsection{Models}
\label{sec:models}

\paragraph{Encoder-based Models.}
Automatic comment classification in \textit{Computational Social Science} (CSS) research—covering tasks such as counter-speech \citep{Bonaldi_Chung_Abercrombie_Guerini_2024}, disagreement detection \citep{De_Kock_Vlachos_2021}, and hate speech analysis \citep{Fortuna_Nunes_2018}—has predominantly relied on \textit{encoder-based transformers}.  
Pretrained models such as BERT \citep{Devlin_Chang_Lee_Toutanova_2019} and RoBERTa \citep{Liu_Ott_Goyal_Du_Joshi_Chen_Levy_Lewis_Zettlemoyer_Stoyanov_2019} are typically fine-tuned on domain-specific datasets via a classification head, enabling effective transfer to specialized applications. 

Following this line of work, we fine-tune the French pre-trained model \textsc{CamemBERT} \citep{Martin_Muller_Suárez_Dupont_Romary_Clergerie_Seddah_Sagot_2020} to detect whether a comment constitutes a \textit{stopping point} ($y{=}1$) or not ($y{=}0$). The model serves as a strong French baseline due to its robust pretraining and established performance in social media tasks. We experimented with both \textsc{CamemBERT} and \textsc{CamemBERT-v2} \citep{Antoun_Kulumba_Touchent_Clergerie_Sagot_Seddah_2024}, and ultimately retained the original \textsc{CamemBERT} model, as it achieved the best performance on our dataset.

The interpretation of a social media comment often depends on its broader publication context, including the post, article, hosting page or group, or parent comment. Encoder-based models—originally designed for sentence-level tasks—tend to overlook this context due to their limited input window and their only later adaptation to social media and conversational data \citep{castelle2018linguistic, Flek_2020}. 
Recent studies have shown that incorporating contextual elements, such as parent comments, post framing, or hosting group community, improves model performances \citep{Park_Mendelsohn_Radhakrishnan_Jain_Kanakagiri_Jurgens_Tsvetkov_2021, bourgeade2024humans, Nouri_Cointet_Clavel_2025}.

We therefore extend our baseline with two contextual variants commonly used in context-aware encoder architectures:  
(i) \textsc{Context Concat}, where contextual text (post message, article title, or a concatenation of all contextual elements) is appended to the input comment using the \texttt{[SEP]} token; and  
(ii) \textsc{Context Embed}, where all contextual elements (post, article title, page/group name, and parent comment) are first concatenated into a single text sequence to produce one context embedding. This embedding is then concatenated with the comment embedding and projected back to a 768-dimensional space before classification.

We define the contextual text as the concatenation of all available elements of the publication context, each preceded by a tag indicating its type (\texttt{[PARENT]} parent comment \texttt{[SEP]} \texttt{[ARTICLE]} article title (in green in Figure~\ref{fig:stop}) \texttt{[SEP]} \texttt{[PAGE]} page name (in orange in Figure~\ref{fig:stop}), etc.). To fit within \textsc{CamemBERT}’s 512-token window, each element is truncated according to predefined contextual limits based on its mean length (see Appendix~\ref{appendix:context} for details). This setup enables a controlled evaluation of how contextual cues contribute to detection performance.

\paragraph{Large Language Models.}
Recent advances in \textit{Large Language Models} (LLMs) have renewed interest in using instruction-tuned models for CSS tasks without task-specific fine-tuning \citep{Mu_Wu_Thorne_Robinson_Aletras_Scarton_Bontcheva_Song_2024}.  
LLMs such as \textsc{Llama-3}, \textsc{Mistral}, and \textsc{Qwen} can perform text classification via prompt adaptation, allowing flexible task formulation through natural language instructions.  
However, as emphasized by \citet{Ziems_Held_Shaikh_Chen_Zhang_Yang_2024}, their performance on social media data—especially for non-English content \citep{pmlr-v239-mohta23a}—remains uneven and often below that of fine-tuned encoders.

To situate stopping point detection within this emerging paradigm, we evaluate a set of state-of-the-art open-weights instruction-tuned LLMs—\textsc{Llama 3.2 3B Instruct} \citep{Grattafiori_Dubey_Jauhri_Pandey_Kadian_Al-Dahle_Letman_Mathur_Schelten_Vaughan_et}, \textsc{Mistral 7B Instruct-v0.2} \citep{Jiang_Sablayrolles_2023}, and \textsc{Qwen2.5 7B Instruct} \citep{Qwen_Yang_Yang_Zhang_Hui_Zheng_Yu_Li_Liu_Huang_et}.  
In addition, we include \textsc{GPT}-4o-\textsc{mini}, a recent proprietary model from OpenAI, as a closed-source reference of comparable size (approximately 8B parameters \citep{zeff2024gpt4omini}). This inclusion enables a direct comparison between open and commercial instruction-tuned models under identical prompting conditions, allowing us to assess the extent to which open-weights LLMs approach the performance of state-of-the-art closed systems on multilingual and noisy social media data.

All models are evaluated under \textit{zero-shot} \citep{Kojima_Gu_Reid_Matsuo_Iwasawa_2022}, \textit{few-shot} \citep{Brown_Mann_Ryder_Subbiah_Kaplan_Dhariwal_Neelakantan_Shyam_Sastry_Askell_2020}, and \textit{chain-of-thought} (CoT) \citep{Wei_Wang_Schuurmans_Bosma_Ichter_Xia_Chi_Le_Zhou_2022} prompting strategies.  
Annotation guidelines are reformulated as natural language prompts following recent practices in CSS research \citep{Ziems_Held_Shaikh_Chen_Zhang_Yang_2024}, and we systematically experiment with prompts in both French and English, as well as with alternative label formulations (\textit{Yes/No}, \textit{1/0}, \textit{Stop/No stop}).  
Consistent with prior findings \citep{Mu_Wu_Thorne_Robinson_Aletras_Scarton_Bontcheva_Song_2024, Ziems_Held_Shaikh_Chen_Zhang_Yang_2024}, we observe that seemingly minor prompt variations can lead to substantial and unpredictable differences in model behavior.  
For the CoT setting, we transform the annotation guidelines into a structured decision tree with illustrative examples, guiding the model through reasoning steps analogous to those followed by human annotators.

As with encoder-based models, we also evaluate the inclusion of publication context—article, post, parent comment, domain, and page or group name—directly within the prompts, using the same tags and truncation limits for comparability.

Overall, we test six prompt configurations: zero-shot and few-shot with and without context, and CoT with context. 
All prompts are available for reference in Appendix~\ref{appendix:prompts}.  
Fine-tuning of LLMs was not attempted due to computational costs limitations; future work may explore fine-tuned or adapter-based approaches.  
All LLM experiments were run with temperature $T{=}0$ for reproducibility.

\subsection{Experiments and Results}

\paragraph{Model Comparison.}
We compare the performance of the model architectures introduced in Section~\ref{sec:models} on the stopping point classification task using the test set. Since stopping points constitute approximately 10\% of the annotated corpus, and the train/validation/test splits were sampled to preserve this class distribution, the random baseline, reported for reference, samples from $P(y{=}1){=}0.1$. All encoder-based models were fine-tuned on the training set following the procedure described in Appendix~\ref{appendix:train}. Table~\ref{tab:main-results} reports the mean $F_{1}$ scores (\%) for all model configurations. For encoder-based models, we include 95\% confidence intervals computed over five independent runs. For LLM-based experiments, the temperature was set to $T=0$ to ensure reproducible outputs; therefore, confidence intervals are not reported for these results. For conciseness, we display results only for the best-performing open-weights model (\textsc{Qwen2.5-7B-Instruct}, as \textsc{Qwen}) and the best-performing proprietary model (\textsc{GPT}-4o-\textsc{mini} as \textsc{GPT}). Results for all evaluated LLMs are provided in Appendix~\ref{appendix:llm-results}. 

\begin{table}[ht!]
\centering
\small
\setlength{\tabcolsep}{4pt}
\begin{tabular}{lc}
\hline
\textbf{Model} & \textbf{$F_{1}$ (\%)} \\
\hline
\textsc{Random} & 16.4 \\
\hline
\textsc{Qwen Zero-Shot (No Context)} & 39.23  \\
\textsc{Qwen Zero-Shot (Context)} & 45.59  \\
\textsc{Qwen Few-Shot (No Context)} & 53.62 \\
\textsc{Qwen Few-Shot (Context)} & 42.52 \\
\textsc{Qwen CoT (Context)} & 45.57 \\
\textsc{GPT Zero-Shot (No Context)} & 53.39  \\
\textsc{GPT Zero-Shot (Context)} & 55.94  \\
\textsc{GPT Few-Shot (No Context)} & 62.94 \\
\textsc{GPT Few-Shot (Context)} & 55.57 \\
\textsc{GPT CoT (Context)} & 54.33 \\
\hline
\textsc{CamemBERT (No Context)} & 74.67 $\pm$ 0.77  \\
\textsc{CamemBERT Context Concat} & 76.96 $\pm$ 1.39 \\
\textsc{CamemBERT Context Embed} &  \textbf{78.09 $\pm$ 0.84} \\
\hline
\end{tabular}
\caption{Mean $F_{1}$ scores (\%) and 95\% confidence intervals over five runs.}
\label{tab:main-results}
\end{table}

Overall, encoder-based models substantially outperform LLMs across all settings, achieving $F_{1}$ scores more than 10 percentage points higher on average. Despite using detailed prompts—explicitly defining the task, providing in-context examples, or even chain-of-thought instructions—LLMs remain far behind the finetuned encoders. This suggests that prompting instruction-tuned LLMs remains insufficient for new, complex, and context-sensitive social media analysis tasks in non-English settings. In particular, as detailed in Section~\ref{sec:annotation}, stopping point detection requires understanding subtle conversational cues and pragmatic shifts that may not be captured without task-specific supervision. This gap aligns with recent findings showing that even instruction-tuned LLMs struggle to generalize to fine-grained, context-dependent classification tasks \citep{Ziems_Held_Shaikh_Chen_Zhang_Yang_2024, Mu_Wu_Thorne_Robinson_Aletras_Scarton_Bontcheva_Song_2024}, especially in non-English settings where pretraining data coverage is lower \citep{pmlr-v239-mohta23a}.
Interestingly, the performance of LLMs varies considerably across prompting strategies. In our experiments, few-shot prompting without context consistently yields the highest $F_{1}$ scores for both \textsc{Qwen} and \textsc{GPT} models, outperforming few-shot, and chain-of-thought with context variants. This suggests that including additional contextual elements or more complex instructions may dilute relevant information or introduce patterns that are too complex to be effectively learned without supervision, preventing LLMs from leveraging the extra context for stopping point detection.

Within the encoder-based models, incorporating conversational context improves performance, confirming prior findings in related social media comment classification tasks such as contextual hate speech and rule-violation detection \citep{Park_Mendelsohn_Radhakrishnan_Jain_Kanakagiri_Jurgens_Tsvetkov_2021, Nouri_Cointet_Clavel_2025}. Among the two integration strategies, the \textsc{Context Embed} model achieves the best performance, outperforming both the context-free and \textsc{Context Concat} variants. This supports the view that embedding the conversational context separately preserves the salience of the comment’s own linguistic features, while concatenation tends to dilute them within a longer input string. 
Together, these results highlight the continued relevance of finetuned, encoder-based architectures for nuanced conversational modeling tasks, where context-integration mechanisms play a decisive role in capturing discourse-level dependencies.

\paragraph{Context Contribution Analysis.}

To better understand which contextual components contribute most to stopping point prediction, we conducted a series of controlled experiments inspired by ablation studies. Since many contextual fields (parent comment or post message) are not available in all instances, we used the \textsc{Context Concat} architecture to maintain consistency across models. In each run, we provided only one contextual element at a time—\textit{article text}, \textit{post message}, \textit{domain name}, \textit{page or group name}, or \textit{parent comment}—alongside the target comment, and evaluated performance under the same conditions as before.

\begin{table}[ht!]
\centering
\small
\setlength{\tabcolsep}{4pt}
\begin{tabular}{lc}
\hline
\textbf{Model Configuration} & \textbf{$F_{1}$ (\%)} \\
\hline
\textsc{No Context} & 74.67 $\pm$ 0.77 \\
\textsc{Parent Comment Concat} & 73.97 $\pm$ 1.09 \\
\textsc{Page/Group Concat} & 76.15 $\pm$ 1.35 \\
\textsc{Domain Concat} & 77.15 $\pm$ 0.65 \\
\textsc{Post Concat} & 77.40 $\pm$ 1.51 \\
\textsc{Article Concat} & 78.03 $\pm$ 1.40 \\
\textsc{Context Concat} & 76.96 $\pm$ 1.39 \\
\textsc{Context Embed} & \textbf{78.09 $\pm$ 0.84} \\
\hline
\end{tabular}
\caption{Performance of fine-tuned \textsc{CamemBERT} classifiers in different context configuration. Mean $F_{1}$ scores (\%) with 95\% confidence intervals over five runs are shown.}
\label{tab:text-concat-results}
\end{table}

Results, displayed in Table~\ref{tab:text-concat-results}, indicate that the \textit{article text} provides the most informative context, yielding the largest performance gain over the no-context baseline. This is consistent with the intuition that stopping points often contain reactions or critiques targeting the shared article. The \textit{post message} also improves classification, as it frequently aligns with or paraphrases the article’s content. Notably, \textit{domain name} and \textit{page or group name} also contribute positively, suggesting that the media source and the social page name carry relevant cues for how ordinary users express criticism in different online contexts. These findings align with \citet{Park_Mendelsohn_Radhakrishnan_Jain_Kanakagiri_Jurgens_Tsvetkov_2021}, who similarly observed that including the community name (in their case, the subreddit) through a \textsc{Context Concat} architecture improved the prediction of moderation rule violations, highlighting the value of contextual information for understanding social media discourse.

\section{Error Analysis}
\label{sec:error_analysis}

To better understand the limitations of our best-performing model (\textsc{Context Embed}), we conducted a manual error analysis on all $360$ misclassified instances from the test set.

Each error was first categorized as a \textit{top-level} comment or a \textit{reply} (see Figure~\ref{fig:replies}), and then annotated as \textit{simple} (decidable from the comment alone) or \textit{complex} (requiring context). Complex cases were further categorized according to the typology described in the Annotation Guidelines (provided in our \href{https://github.com/celia-nouri/SPOT-benchmark}{code repository}) (explicit markers, reported speech, reply-dependency, irony, short fragments, multi-turn phenomena). Categories are not mutually exclusive, as a single instance may belong to multiple sources of ambiguity.

\paragraph{False positives analysis.} 

Among the 203 false positives, 85.2\% are labelled \emph{complex}. Many stem from comments that contain explicit critical markers (77.3\%) while reacting to posts that report or quote third-party claims (30.0\%). In these situations the user’s reaction typically attacks the quoted claim or source rather than performing a critical intervention on the act of reposting. For example, in response to a post claiming \textit{“President Erdogan encourages a Turkish girl to die as a martyr”}, the comment \textit{“Nonsense! He should drop the mic and go die as a martyr himself”} is an emotional reaction aimed at the Erdogan and his rhetoric. It expresses outrage but does not invite readers to verify or report the post, and so should not be annotated as a stopping point. Reply-specific phenomena also matter: 36.5\% of errors involve reply dynamics and 11.8\% match a ``reply to a stopping point'' pattern in which the model mistakes a reactive turn for a critical intervention. Additionally, irony accounts for 16.3\% of false positive cases, and short comments for 5.4\%. Together, these observations reveal a clear pattern: the context-aware encoder still over-weights surface cues (``fake'', ``montage'', URLs, numerical claims) and contextual signals associated with controversy, which leads it to misread phatic, ironic or meta-discursive reactions as stopping points.

\paragraph{False negative analysis.}

Out of 157 false negatives, 79\% correspond to complex cases, confirming that most model errors occur in linguistically or contextually ambiguous situations. Replies represent 45\% of false negatives, compared to 34.3\% in the overall dataset, suggesting that replies constitute a particularly challenging structure. Among these, 38\% are replies to stopping points, a type of interaction that is also difficult for human annotators. The most frequent source of false negative errors is the absence of explicit refutation markers (51.6\%), showing that the model tends to rely on surface lexical cues rather than pragmatic or discourse-level information. Irony and humor (12.7\%) also account for a notable share of errors, illustrating the limitations of encoder models in capturing implicit stance or socio-cultural nuances.

\begin{table}[ht!]
\centering
\small
\setlength{\tabcolsep}{4pt}
\begin{tabular}{lcc}
\hline
\textbf{Category} & \textbf{Count} & \textbf{Percent (\%)} \\
\hline
\multicolumn{3}{l}{\textbf{False Positives (N = 203)}} \\
\hline
No refutation with markers & 157 & 77.3 \\
Reported speech & 61 & 30.0 \\
Irony / humor & 33 & 16.3 \\
Reply to a stopping point & 24 & 11.8 \\
Short or fragmentary & 11 & 5.4 \\
\hline
\multicolumn{3}{l}{\textbf{False Negatives (N = 157)}} \\
\hline
Refutation without markers & 81 & 51.6 \\
Reply to a stopping point & 27 & 17.2 \\
Irony or humor & 20 & 12.7 \\
\hline
\end{tabular}
\caption{Error types for the \textsc{Context Embed} model. Percentages are within each class (FP/FN); only categories with $>$10 cases are shown.}
\label{tab:error-analysis}
\end{table}

In summary, the error analysis shows that context-aware encoders over-rely on lexical markers, producing false positives when explicit cues of criticism appear in supportive comments, and false negatives when criticism is implicit or ironic. Stopping points are also detected more reliably in top-level comments than in replies, highlighting the need to model conversational structure or use separate models for different comment types.

\section{Conclusion and Future Directions}
We present \textsc{SPOT}, the first large-scale corpus of 43{,}305 French Facebook comments manually annotated to capture \emph{stopping points}. \textsc{SPOT} extends beyond conventional notions of user corrections or fact-checking, revealing diverse everyday critical interventions that pause, question, or redirect online discourse. The corpus includes detailed contextual metadata (post, article, domain, page or group, and parent comment) and is accompanied by comprehensive annotation guidelines to ensure transparency and reproducibility.  

Using \textsc{SPOT}, we benchmarked fine-tuned encoders and instruction-tuned LLMs across multiple prompting strategies. Supervised encoders outperform prompted LLMs by over 10 $F_{1}$ points, showing that for nuanced phenomena like stopping points—where meaning depends on social and contextual cues—models benefit more from explicit supervision than from general-purpose instructions, particularly in non-English settings. Adding contextual metadata further improves $F_{1}$ scores from 0.75 to 0.78, emphasizing the importance of analyzing comments within their broader publication context. Error analysis shows that encoders still struggle when lexical markers contradict intent, such as irony or emotionally charged supportive comments, indicating overreliance on surface signals. Stopping points are detected more reliably in top-level comments than in replies, highlighting the need to model conversational structure or design separate models for different comment types.

Future work will improve encoder architectures to better capture conversational and social context, moving beyond linear concatenation toward graph-based or hierarchical models and integrating multimodal signals (images, videos). We will also extend the current binary formulation to a multi-label classification task that can automatically distinguish different types of stopping points, and we will separate detection of top-level comments from nested replies (e.g., distinct models or pipelines) to better capture hierarchical conversational dynamics. Finally, we plan to study critical interventions across platforms (Reddit, YouTube) and languages, while iteratively refining our annotation guide to enhance label quality and model robustness. Collectively, these directions aim to advance computational sociology and NLP by modeling online interventions as socially situated, context-dependent phenomena.

\section{Ethical Considerations and Limitations}

The SPOT corpus contains user-generated content from public Facebook pages and groups, including comments on posts flagged as potentially misleading. Although the posts and comments were public at the time of collection, some content may later be deleted or restricted by its authors or communities. Additionally, the dataset includes user reactions in potentially sensitive contexts, such as disagreements, critiques, or emotionally charged responses.

To protect user privacy and follow established guidelines for ethical social media research \citep{townsend2016social}, we applied several precautions: (i) all user identifiers and profile names were anonymized; and (ii) we do not distribute the dataset publicly to avoid preserving or republishing sensitive material. 

Access to SPOT is granted only for academic research upon request through a secure institutional repository. Each request is evaluated to ensure that the proposed use aligns with ethical guidelines and that the data will be handled responsibly. This controlled-access model balances reproducibility and research transparency with the protection of individual privacy and community norms.

\section{Acknowledgements}

We would like to thank Dominique Cardon, Salim Hafid, Sofia Imbert de Trémiolles, Aina Garí Soler, Théophile Pénigaud de Mourgues, Paul Lerner, and Carlo Romano Marcello Alessandro Santagiustina for their careful reading of the manuscript and for their thoughtful comments and suggestions.

This work was partially funded by the "AI For Democracy Democratic Commons" project (Bpifrance's 'Digital Commons for Generative AI', France 2030), and the French National Research Agency (ANR) under the SINNet project (ANR-23-CE23-0033-01), the France 2030 program PRAIRIE (ANR-23-IACL-0008), PostGenAI@Paris (ANR-23-IACL-0007), and TIERED (ANR-22-EXES-0014).

\section*{References}\label{sec:reference}

\bibliographystyle{lrec2026-natbib}
\bibliography{lrec2026-example}

@article{kuccuk2020stance, title={Stance Detection: A Survey}, volume={53}, ISSN={0360-0300}, DOI={10.1145/3369026}, abstractNote={Automatic elicitation of semantic information from natural language texts is an important research problem with many practical application areas. Especially after the recent proliferation of online content through channels such as social media sites, news portals, and forums; solutions to problems such as sentiment analysis, sarcasm/controversy/veracity/rumour/fake news detection, and argument mining gained increasing impact and significance, revealed with large volumes of related scientific publications. In this article, we tackle an important problem from the same family and present a survey of stance detection in social media posts and (online) regular texts. Although stance detection is defined in different ways in different application settings, the most common definition is “automatic classification of the stance of the producer of a piece of text, towards a target, into one of these three classes: {Favor, Against, Neither}.” Our survey includes definitions of related problems and concepts, classifications of the proposed approaches so far, descriptions of the relevant datasets and tools, and related outstanding issues. Stance detection is a recent natural language processing topic with diverse application areas, and our survey article on this newly emerging topic will act as a significant resource for interested researchers and practitioners.}, number={1}, journal={ACM Comput. Surv.}, author={Küçük, Dilek and Can, Fazli}, year={2020}, month=feb, pages={12:1-12:37} }

@article{ziegele2020lonely, address={US}, title={Lonely together? Identifying the determinants of collective corrective action against uncivil comments}, volume={22}, ISSN={1461-7315}, DOI={10.1177/1461444819870130}, abstractNote={Journalists, scholars, and citizens have raised concerns regarding the high share of incivility in comment sections of news outlets. The current study surveyed members of the social movement #ichbinhier, which aims at collectively countering uncivil comments to cultivate a civil discussion atmosphere in comment sections. We root the activities of #ichbinhier as corrective action and identify the determinants of the members’ engagement by integrating research on bystander behavior and collective action. The findings of our survey show that factors pertaining to individual skills, perceived responsibility, and expected benefits relate to the members’ likelihood to engage against uncivil online comments. Regarding factors derived from collective action research, group efficacy and knowledge of the rules and structures of the movement account for higher levels of engagement. These results shed light on the factors that motivate and inhibit #ichbinhier members—and, potentially, Facebook users in general—to engage against uncivil comments. (PsycInfo Database Record (c) 2021 APA, all rights reserved)}, number={5}, journal={New Media \& Society}, publisher={Sage Publications}, author={Ziegele, Marc and Naab, Teresa K. and Jost, Pablo}, year={2020}, pages={731–751} }

@article{watson2019will, address={US}, title={Who will intervene to save news comments? Deviance and social control in communities of news commenters}, volume={21}, ISSN={1461-7315}, DOI={10.1177/1461444819828328}, abstractNote={Which bystanders will confront racist, misogynist, personal attacks in news comment sections? This article applies sociological concepts of deviance and social control to categorize efforts to moderate online news comments. Three dimensions of social control are theorized: affirming and sanctioning social control, formal and informal social control, and direct and indirect social control. Particular focus is on indirect informal social control (i.e. rating and reporting of news comments) in order to examine which users are likely to intervene to maintain social order. An analysis of secondary data from a survey of online news users found that demographics play an important role—younger, wealthier, White, males are most likely to report abusive comments. Trust in the news media and authoritarian personality traits also significantly predicted bystander intervention. Theoretical implications for the role of social control in enforcing social norms in news comment spaces and for professional comment moderation are discussed. (PsycINFO Database Record (c) 2019 APA, all rights reserved)}, number={8}, journal={New Media \& Society}, publisher={Sage Publications}, author={Watson, Brendan R. and Peng, Zhao and Lewis, Seth C.}, year={2019}, pages={1840–1858} }

@article{porten2020online, title={Online Civic Intervention: A New Form of Political Participation Under Conditions of a Disruptive Online Discourse}, volume={14}, ISSN={1932-8036}, abstractNote={In the everyday practice of online communication, we observe users deliberately reporting abusive content or opposing hate speech through counterspeech, while at the same time, online platforms are increasingly relying on and supporting this kind of user action to fight disruptive online behavior. We refer to this type of user engagement as online civic intervention (OCI) and regard it as a new form of user-based political participation in the digital sphere that contributes to an accessible and reasoned public discourse. Because OCI has received little scholarly attention thus far, this article conceptualizes low- and high-threshold types of OCI as different kinds of user responses to common disruptive online behavior such as hate speech or hostility toward the media. Against the background of participation research, we propose a theoretically grounded individual-level model that serves to explain OCI.}, journal={International Journal of Communication}, author={Porten-Cheé, Pablo and Kunst, Marlene and Emmer, Martin}, year={2020}, month=jan, pages={21–21}, language={en} }

@article{bode2024user, title={User correction}, volume={56}, ISSN={2352-250X}, DOI={10.1016/j.copsyc.2023.101786}, abstractNote={This paper reviews the existing literature on user correction to consider its value for combating misinformation on social media. We discuss the effectiveness of user correction in reducing misperceptions, and synthesize best practices, highlighting the dual audiences for public correction on social media. We outline how often user correction occurs across contexts, countries, and social media platforms. We pay special attention to the methodological constraints in existing research, emphasizing the need for using diverse and interdisciplinary methods, including longitudinal surveys and experiments, computational methods, realistic simulated environments, and qualitative methods. We call for a more comprehensive understanding of user correction in terms of its long-term and downstream effects on social media platforms.}, journal={Current Opinion in Psychology}, author={Bode, Leticia and Vraga, Emily K. and Tang, Rongwei}, year={2024}, month=apr, pages={101786} }

@article{buerger2021, address={Rochester, NY}, type={SSRN Scholarly Paper}, title={Counterspeech: A Literature Review}, url={https://papers.ssrn.com/abstract=3829816}, DOI={10.2139/ssrn.3829816}, abstractNote={Every day, internet users encounter hateful and dangerous speech online, and some of them choose to respond directly in order to refute or undermine it. We call this counterspeech. Many of those who have taken on this volunteer effort go about it alone, while others organize into groups to coordinate responses and support each other. Some staff at social platforms have touted counterspeech as a method of reducing online hate, but, like Justice Louis Brandeis, who also opined that the remedy for bad speech is good speech, they don’t cite any evidence for their assertions. This is an effort to bridge that gap by answering what should be a prominent question: what does the scholarly literature have to say about the effectiveness of counterspeech?}, number={3829816}, publisher={Social Science Research Network}, author={Buerger, Catherine and Wright, Lucas}, year={2019}, month=nov, language={en} }

@article{vosoughi2018spread, title={The spread of true and false news online}, volume={359}, DOI={10.1126/science.aap9559}, abstractNote={We investigated the differential diffusion of all of the verified true and false news stories distributed on Twitter from 2006 to 2017. The data comprise ~126,000 stories tweeted by ~3 million people more than 4.5 million times. We classified news as true or false using information from six independent fact-checking organizations that exhibited 95 to 98% agreement on the classifications. Falsehood diffused significantly farther, faster, deeper, and more broadly than the truth in all categories of information, and the effects were more pronounced for false political news than for false news about terrorism, natural disasters, science, urban legends, or financial information. We found that false news was more novel than true news, which suggests that people were more likely to share novel information. Whereas false stories inspired fear, disgust, and surprise in replies, true stories inspired anticipation, sadness, joy, and trust. Contrary to conventional wisdom, robots accelerated the spread of true and false news at the same rate, implying that false news spreads more than the truth because humans, not robots, are more likely to spread it.}, number={6380}, journal={Science}, publisher={American Association for the Advancement of Science}, author={Vosoughi, Soroush and Roy, Deb and Aral, Sinan}, year={2018}, month=mar, pages={1146–1151} }

@article{delvicario2016spreading,
author = {Michela Del Vicario  and Alessandro Bessi  and Fabiana Zollo  and Fabio Petroni  and Antonio Scala  and Guido Caldarelli  and H. Eugene Stanley  and Walter Quattrociocchi },
title = {The spreading of misinformation online},
journal = {Proceedings of the National Academy of Sciences},
volume = {113},
number = {3},
pages = {554-559},
year = {2016},
doi = {10.1073/pnas.1517441113},
URL = {https://www.pnas.org/doi/abs/10.1073/pnas.1517441113},
eprint = {https://www.pnas.org/doi/pdf/10.1073/pnas.1517441113}}

@phdthesis{berriche:tel-05409923,
  TITLE = {{Tu crois que c'est vrai ? : diversit{\'e} des r{\'e}gimes d'{\'e}nonciation face aux fake news et m{\'e}canismes d'autor{\'e}gulation conversationnelle}},
  AUTHOR = {Berriche, Manon},
  URL = {https://theses.hal.science/tel-05409923},
  NUMBER = {2024UNIP7302},
  SCHOOL = {{Universit{\'e} Paris Cit{\'e}}},
  YEAR = {2024},
  MONTH = Dec,
  TYPE = {Theses},
  HAL_ID = {tel-05409923},
  HAL_VERSION = {v1},
}

@article{Liu_Ott_Goyal_Du_Joshi_Chen_Levy_Lewis_Zettlemoyer_Stoyanov_2019,
  author       = {Yinhan Liu and
                  Myle Ott and
                  Naman Goyal and
                  Jingfei Du and
                  Mandar Joshi and
                  Danqi Chen and
                  Omer Levy and
                  Mike Lewis and
                  Luke Zettlemoyer and
                  Veselin Stoyanov},
  title        = {RoBERTa: {A} Robustly Optimized {BERT} Pretraining Approach},
  journal      = {CoRR},
  volume       = {abs/1907.11692},
  year         = {2019},
  url          = {http://arxiv.org/abs/1907.11692},
  eprinttype    = {arXiv},
  eprint       = {1907.11692},
  bibsource    = {dblp computer science bibliography, https://dblp.org}
}

@misc{Messing_DeGregorio_Hillenbrand_King_Mahanti_Mukerjee_Nayak_Persily_State_Wilkins_2023, title={Facebook Privacy-Protected Full URLs Data Set}, rights={http://creativecommons.org/publicdomain/zero/1.0}, url={https://dataverse.harvard.edu/dataset.xhtml?persistentId=doi:10.7910/DVN/TDOAPG}, DOI={10.7910/DVN/TDOAPG}, abstractNote={This is a codebook for data on the demographics of people who viewed, shared, and otherwise interacted with web pages (URLs) shared on Facebook, between January 1, 2017 and October 31, 2022. The data has about 68 million URLs, over 3.1 trillion rows, and over 71 trillion cell values. It results from a collaboration between Facebook and Social Science One (at IQSS at Harvard), originally prepared for Social Science One grantees and describes the “full” URLs dataset, including its scope, structure, and fields. This is version 10 of the codebook and data (released 4/13/2023), first described by Gary King and Nathaniel Persily at https://socialscience.one/blog/update-social-science-one.}, publisher={Harvard Dataverse}, author={Messing, Solomon and DeGregorio, Christina and Hillenbrand, Bennett and King, Gary and Mahanti, Saurav and Mukerjee, Zagreb and Nayak, Chaya and Persily, Nate and State, Bogdan and Wilkins, Arjun}, year={2023}, month=apr, language={fr} }

@inproceedings{Martin_Muller_Suárez_Dupont_Romary_Clergerie_Seddah_Sagot_2020, title={CamemBERT: a Tasty French Language Model}, url={http://arxiv.org/abs/1911.03894}, DOI={10.18653/v1/2020.acl-main.645}, abstractNote={Pretrained language models are now ubiquitous in Natural Language Processing. Despite their success, most available models have either been trained on English data or on the concatenation of data in multiple languages. This makes practical use of such models --in all languages except English-- very limited. In this paper, we investigate the feasibility of training monolingual Transformer-based language models for other languages, taking French as an example and evaluating our language models on part-of-speech tagging, dependency parsing, named entity recognition and natural language inference tasks. We show that the use of web crawled data is preferable to the use of Wikipedia data. More surprisingly, we show that a relatively small web crawled dataset (4GB) leads to results that are as good as those obtained using larger datasets (130+GB). Our best performing model CamemBERT reaches or improves the state of the art in all four downstream tasks.}, note={arXiv:1911.03894 [cs]}, booktitle={Proceedings of the 58th Annual Meeting of the Association for Computational Linguistics}, author={Martin, Louis and Muller, Benjamin and Suárez, Pedro Javier Ortiz and Dupont, Yoann and Romary, Laurent and Clergerie, Éric Villemonte de la and Seddah, Djamé and Sagot, Benoît}, year={2020}, pages={7203–7219} }

@article{Antoun_Kulumba_Touchent_Clergerie_Sagot_Seddah_2024, title={CamemBERT 2.0: A Smarter French Language Model Aged to Perfection}, url={http://arxiv.org/abs/2411.08868}, DOI={10.48550/arXiv.2411.08868}, abstractNote={French language models, such as CamemBERT, have been widely adopted across industries for natural language processing (NLP) tasks, with models like CamemBERT seeing over 4 million downloads per month. However, these models face challenges due to temporal concept drift, where outdated training data leads to a decline in performance, especially when encountering new topics and terminology. This issue emphasizes the need for updated models that reflect current linguistic trends. In this paper, we introduce two new versions of the CamemBERT base model-CamemBERTav2 and CamemBERTv2-designed to address these challenges. CamemBERTav2 is based on the DeBERTaV3 architecture and makes use of the Replaced Token Detection (RTD) objective for better contextual understanding, while CamemBERTv2 is built on RoBERTa, which uses the Masked Language Modeling (MLM) objective. Both models are trained on a significantly larger and more recent dataset with longer context length and an updated tokenizer that enhances tokenization performance for French. We evaluate the performance of these models on both general-domain NLP tasks and domain-specific applications, such as medical field tasks, demonstrating their versatility and effectiveness across a range of use cases. Our results show that these updated models vastly outperform their predecessors, making them valuable tools for modern NLP systems. All our new models, as well as intermediate checkpoints, are made openly available on Huggingface.}, note={arXiv:2411.08868 [cs]}, number={arXiv:2411.08868}, publisher={arXiv}, author={Antoun, Wissam and Kulumba, Francis and Touchent, Rian and Clergerie, Éric de la and Sagot, Benoît and Seddah, Djamé}, year={2024}, month=nov }

@inproceedings{bourgeade2024humans,
    title = "Humans Need Context, What about Machines? Investigating Conversational Context in Abusive Language Detection",
    author = "Bourgeade, Tom  and
      Li, Zongmin  and
      Benamara, Farah  and
      Moriceau, V{\'e}ronique  and
      Su, Jian  and
      Sun, Aixin",
    editor = "Calzolari, Nicoletta  and
      Kan, Min-Yen  and
      Hoste, Veronique  and
      Lenci, Alessandro  and
      Sakti, Sakriani  and
      Xue, Nianwen",
    booktitle = "Proceedings of the 2024 Joint International Conference on Computational Linguistics, Language Resources and Evaluation (LREC-COLING 2024)",
    month = may,
    year = "2024",
    address = "Torino, Italia",
    publisher = "ELRA and ICCL",
    url = "https://aclanthology.org/2024.lrec-main.740",
    pages = "8438--8452",
}

@inproceedings{castelle2018linguistic,
  title={The Linguistic Ideologies of Deep Abusive Language Classification},
  author={Castelle, Michael},
  booktitle={Proceedings of the Workshop on Abusive Language Online (ALW)},
  year={2018},
  doi={10.18653/v1/W18-5120},
  corpusid={53644609},
  publisher={Association for Computational Linguistics},
  address={Melbourne, Australia},
  pages={160--170}
}

@inproceedings{Nouri_Cointet_Clavel_2025, address={Vienna, Austria}, title={Graphically Speaking: Unmasking Abuse in Social Media with Conversation Insights}, ISBN={979-8-89176-251-0}, url={https://aclanthology.org/2025.acl-long.894/}, DOI={10.18653/v1/2025.acl-long.894}, abstractNote={Detecting abusive language in social media conversations poses significant challenges, as identifying abusiveness often depends on the conversational context, characterized by the content and topology of preceding comments. Traditional Abusive Language Detection (ALD) models often overlook this context, which can lead to unreliable performance metrics. Recent Natural Language Processing (NLP) approaches that incorporate conversational context often rely on limited or overly simplified representations of this context, leading to inconsistent and sometimes inconclusive results. In this paper, we propose a novel approach that utilizes graph neural networks (GNNs) to model social media conversations as graphs, where nodes represent comments, and edges capture reply structures. We systematically investigate various graph representations and context windows to identify the optimal configurations for ALD. Our GNN model outperforms both context-agnostic baselines and linear context-aware methods, achieving significant improvements in F1 scores. These findings demonstrate the critical role of structured conversational context and establish GNNs as a robust framework for advancing context-aware ALD.}, booktitle={Proceedings of the 63rd Annual Meeting of the Association for Computational Linguistics (Volume 1: Long Papers)}, publisher={Association for Computational Linguistics}, author={Nouri, Célia and Cointet, Jean-Philippe and Clavel, Chloé}, editor={Che, Wanxiang and Nabende, Joyce and Shutova, Ekaterina and Pilehvar, Mohammad Taher}, year={2025}, month=july, pages={18271–18286} }

@inproceedings{Waseem_2016, address={Austin, Texas}, title={Are You a Racist or Am I Seeing Things? Annotator Influence on Hate Speech Detection on Twitter}, url={https://aclanthology.org/W16-5618/}, DOI={10.18653/v1/W16-5618}, booktitle={Proceedings of the First Workshop on NLP and Computational Social Science}, publisher={Association for Computational Linguistics}, author={Waseem, Zeerak}, editor={Bamman, David and Doğruöz, A. Seza and Eisenstein, Jacob and Hovy, Dirk and Jurgens, David and O’Connor, Brendan and Oh, Alice and Tsur, Oren and Volkova, Svitlana}, year={2016}, month=nov, pages={138–142} }

@inproceedings{Devlin_Chang_Lee_Toutanova_2019, address={Minneapolis, Minnesota}, title={BERT: Pre-training of Deep Bidirectional Transformers for Language Understanding}, url={https://aclanthology.org/N19-1423/}, DOI={10.18653/v1/N19-1423}, abstractNote={We introduce a new language representation model called BERT, which stands for Bidirectional Encoder Representations from Transformers. Unlike recent language representation models (Peters et al., 2018a; Radford et al., 2018), BERT is designed to pre-train deep bidirectional representations from unlabeled text by jointly conditioning on both left and right context in all layers. As a result, the pre-trained BERT model can be fine-tuned with just one additional output layer to create state-of-the-art models for a wide range of tasks, such as question answering and language inference, without substantial task-specific architecture modifications. BERT is conceptually simple and empirically powerful. It obtains new state-of-the-art results on eleven natural language processing tasks, including pushing the GLUE score to 80.5 (7.7 point absolute improvement), MultiNLI accuracy to 86.7% (4.6% absolute improvement), SQuAD v1.1 question answering Test F1 to 93.2 (1.5 point absolute improvement) and SQuAD v2.0 Test F1 to 83.1 (5.1 point absolute improvement).}, booktitle={Proceedings of the 2019 Conference of the North American Chapter of the Association for Computational Linguistics: Human Language Technologies, Volume 1 (Long and Short Papers)}, publisher={Association for Computational Linguistics}, author={Devlin, Jacob and Chang, Ming-Wei and Lee, Kenton and Toutanova, Kristina}, editor={Burstein, Jill and Doran, Christy and Solorio, Thamar}, year={2019}, month=june, pages={4171–4186} }

@inproceedings{Flek_2020, address={Online}, title={Returning the N to NLP: Towards Contextually Personalized Classification Models}, url={https://aclanthology.org/2020.acl-main.700/}, DOI={10.18653/v1/2020.acl-main.700}, abstractNote={Most NLP models today treat language as universal, even though socio- and psycholingustic research shows that the communicated message is influenced by the characteristics of the speaker as well as the target audience. This paper surveys the landscape of personalization in natural language processing and related fields, and offers a path forward to mitigate the decades of deviation of the NLP tools from sociolingustic findings, allowing to flexibly process the “natural” language of each user rather than enforcing a uniform NLP treatment. It outlines a possible direction to incorporate these aspects into neural NLP models by means of socially contextual personalization, and proposes to shift the focus of our evaluation strategies accordingly.}, booktitle={Proceedings of the 58th Annual Meeting of the Association for Computational Linguistics}, publisher={Association for Computational Linguistics}, author={Flek, Lucie}, editor={Jurafsky, Dan and Chai, Joyce and Schluter, Natalie and Tetreault, Joel}, year={2020}, month=july, pages={7828–7838} }

@inproceedings{Park_Mendelsohn_Radhakrishnan_Jain_Kanakagiri_Jurgens_Tsvetkov_2021, address={Punta Cana, Dominican Republic}, title={Detecting Community Sensitive Norm Violations in Online Conversations}, url={https://aclanthology.org/2021.findings-emnlp.288/}, DOI={10.18653/v1/2021.findings-emnlp.288}, abstractNote={Online platforms and communities establish their own norms that govern what behavior is acceptable within the community. Substantial effort in NLP has focused on identifying unacceptable behaviors and, recently, on forecasting them before they occur. However, these efforts have largely focused on toxicity as the sole form of community norm violation. Such focus has overlooked the much larger set of rules that moderators enforce. Here, we introduce a new dataset focusing on a more complete spectrum of community norms and their violations in the local conversational and global community contexts. We introduce a series of models that use this data to develop context- and community-sensitive norm violation detection, showing that these changes give high performance.}, booktitle={Findings of the Association for Computational Linguistics: EMNLP 2021}, publisher={Association for Computational Linguistics}, author={Park, Chan Young and Mendelsohn, Julia and Radhakrishnan, Karthik and Jain, Kinjal and Kanakagiri, Tushar and Jurgens, David and Tsvetkov, Yulia}, editor={Moens, Marie-Francine and Huang, Xuanjing and Specia, Lucia and Yih, Scott Wen-tau}, year={2021}, month=nov, pages={3386–3397} }

@article{Ziems_Held_Shaikh_Chen_Zhang_Yang_2024, address={Cambridge, MA}, title={Can Large Language Models Transform Computational Social Science?}, volume={50}, DOI={10.1162/coli_a_00502}, abstractNote={Large language models (LLMs) are capable of successfully performing many language processing tasks zero-shot (without training data). If zero-shot LLMs can also reliably classify and explain social phenomena like persuasiveness and political ideology, then LLMs could augment the computational social science (CSS) pipeline in important ways. This work provides a road map for using LLMs as CSS tools. Towards this end, we contribute a set of prompting best practices and an extensive evaluation pipeline to measure the zero-shot performance of 13 language models on 25 representative English CSS benchmarks. On taxonomic labeling tasks (classification), LLMs fail to outperform the best fine-tuned models but still achieve fair levels of agreement with humans. On free-form coding tasks (generation), LLMs produce explanations that often exceed the quality of crowdworkers’ gold references. We conclude that the performance of today’s LLMs can augment the CSS research pipeline in two ways: (1) serving as zero-shot data annotators on human annotation teams, and (2) bootstrapping challenging creative generation tasks (e.g., explaining the underlying attributes of a text). In summary, LLMs are posed to meaningfully participate in social science analysis in partnership with humans.}, number={1}, journal={Computational Linguistics}, publisher={MIT Press}, author={Ziems, Caleb and Held, William and Shaikh, Omar and Chen, Jiaao and Zhang, Zhehao and Yang, Diyi}, year={2024}, month=mar, pages={237–291} }

@inproceedings{Bonaldi_Chung_Abercrombie_Guerini_2024, address={Mexico City, Mexico}, title={NLP for Counterspeech against Hate: A Survey and How-To Guide}, url={https://aclanthology.org/2024.findings-naacl.221/}, DOI={10.18653/v1/2024.findings-naacl.221}, abstractNote={In recent years, counterspeech has emerged as one of the most promising strategies to fight online hate. These non-escalatory responses tackle online abuse while preserving the freedom of speech of the users, and can have a tangible impact in reducing online and offline violence. Recently, there has been growing interest from the Natural Language Processing (NLP) community in addressing the challenges of analysing, collecting, classifying, and automatically generating counterspeech, to reduce the huge burden of manually producing it. In particular, researchers have taken different directions in addressing these challenges, thus providing a variety of related tasks and resources. In this paper, we provide a guide for doing research on counterspeech, by describing - with detailed examples - the steps to undertake, and providing best practices that can be learnt from the NLP studies on this topic. Finally, we discuss open challenges and future directions of counterspeech research in NLP.}, booktitle={Findings of the Association for Computational Linguistics: NAACL 2024}, publisher={Association for Computational Linguistics}, author={Bonaldi, Helena and Chung, Yi-Ling and Abercrombie, Gavin and Guerini, Marco}, editor={Duh, Kevin and Gomez, Helena and Bethard, Steven}, year={2024}, month=june, pages={3480–3499} }

@article{Fortuna_Nunes_2018, title={A Survey on Automatic Detection of Hate Speech in Text}, volume={51}, ISSN={0360-0300}, DOI={10.1145/3232676}, abstractNote={The scientific study of hate speech, from a computer science point of view, is recent. This survey organizes and describes the current state of the field, providing a structured overview of previous approaches, including core algorithms, methods, and main features used. This work also discusses the complexity of the concept of hate speech, defined in many platforms and contexts, and provides a unifying definition. This area has an unquestionable potential for societal impact, particularly in online communities and digital media platforms. The development and systematization of shared resources, such as guidelines, annotated datasets in multiple languages, and algorithms, is a crucial step in advancing the automatic detection of hate speech.}, number={4}, journal={ACM Comput. Surv.}, author={Fortuna, Paula and Nunes, Sérgio}, year={2018}, month=july, pages={85:1-85:30} }

@article{Mu_Wu_Thorne_Robinson_Aletras_Scarton_Bontcheva_Song_2024, title={Navigating Prompt Complexity for Zero-Shot Classification: A Study of Large Language Models in Computational Social Science}, url={http://arxiv.org/abs/2305.14310}, DOI={10.48550/arXiv.2305.14310}, abstractNote={Instruction-tuned Large Language Models (LLMs) have exhibited impressive language understanding and the capacity to generate responses that follow specific prompts. However, due to the computational demands associated with training these models, their applications often adopt a zero-shot setting. In this paper, we evaluate the zero-shot performance of two publicly accessible LLMs, ChatGPT and OpenAssistant, in the context of six Computational Social Science classification tasks, while also investigating the effects of various prompting strategies. Our experiments investigate the impact of prompt complexity, including the effect of incorporating label definitions into the prompt; use of synonyms for label names; and the influence of integrating past memories during foundation model training. The findings indicate that in a zero-shot setting, current LLMs are unable to match the performance of smaller, fine-tuned baseline transformer models (such as BERT-large). Additionally, we find that different prompting strategies can significantly affect classification accuracy, with variations in accuracy and F1 scores exceeding 10%.}, note={arXiv:2305.14310 [cs]}, number={arXiv:2305.14310}, publisher={arXiv}, author={Mu, Yida and Wu, Ben P. and Thorne, William and Robinson, Ambrose and Aletras, Nikolaos and Scarton, Carolina and Bontcheva, Kalina and Song, Xingyi}, year={2024}, month=mar }

@article{Jiang_Sablayrolles_2023, title={Mistral 7B}, url={http://arxiv.org/abs/2310.06825}, DOI={10.48550/arXiv.2310.06825}, abstractNote={We introduce Mistral 7B v0.1, a 7-billion-parameter language model engineered for superior performance and efficiency. Mistral 7B outperforms Llama 2 13B across all evaluated benchmarks, and Llama 1 34B in reasoning, mathematics, and code generation. Our model leverages grouped-query attention (GQA) for faster inference, coupled with sliding window attention (SWA) to effectively handle sequences of arbitrary length with a reduced inference cost. We also provide a model fine-tuned to follow instructions, Mistral 7B -- Instruct, that surpasses the Llama 2 13B -- Chat model both on human and automated benchmarks. Our models are released under the Apache 2.0 license.}, note={arXiv:2310.06825 [cs]}, number={arXiv:2310.06825}, publisher={arXiv}, author={Jiang, Albert Q. and Sablayrolles, Alexandre and Mensch, Arthur and Bamford, Chris and Chaplot, Devendra Singh and Casas, Diego de las and Bressand, Florian and Lengyel, Gianna and Lample, Guillaume and Saulnier, Lucile and Lavaud, Lélio Renard and Lachaux, Marie-Anne and Stock, Pierre and Scao, Teven Le and Lavril, Thibaut and Wang, Thomas and Lacroix, Timothée and Sayed, William El}, year={2023}, month=oct }

@article{Qwen_Yang_Yang_Zhang_Hui_Zheng_Yu_Li_Liu_Huang_et, title={Qwen2.5 Technical Report}, url={http://arxiv.org/abs/2412.15115}, DOI={10.48550/arXiv.2412.15115}, abstractNote={In this report, we introduce Qwen2.5, a comprehensive series of large language models (LLMs) designed to meet diverse needs. Compared to previous iterations, Qwen 2.5 has been significantly improved during both the pre-training and post-training stages. In terms of pre-training, we have scaled the high-quality pre-training datasets from the previous 7 trillion tokens to 18 trillion tokens. This provides a strong foundation for common sense, expert knowledge, and reasoning capabilities. In terms of post-training, we implement intricate supervised finetuning with over 1 million samples, as well as multistage reinforcement learning. Post-training techniques enhance human preference, and notably improve long text generation, structural data analysis, and instruction following. To handle diverse and varied use cases effectively, we present Qwen2.5 LLM series in rich sizes. Open-weight offerings include base and instruction-tuned models, with quantized versions available. In addition, for hosted solutions, the proprietary models currently include two mixture-of-experts (MoE) variants: Qwen2.5-Turbo and Qwen2.5-Plus, both available from Alibaba Cloud Model Studio. Qwen2.5 has demonstrated top-tier performance on a wide range of benchmarks evaluating language understanding, reasoning, mathematics, coding, human preference alignment, etc. Specifically, the open-weight flagship Qwen2.5-72B-Instruct outperforms a number of open and proprietary models and demonstrates competitive performance to the state-of-the-art open-weight model, Llama-3-405B-Instruct, which is around 5 times larger. Qwen2.5-Turbo and Qwen2.5-Plus offer superior cost-effectiveness while performing competitively against GPT-4o-mini and GPT-4o respectively. Additionally, as the foundation, Qwen2.5 models have been instrumental in training specialized models such as Qwen2.5-Math, Qwen2.5-Coder, QwQ, and multimodal models.}, note={arXiv:2412.15115 [cs]}, number={arXiv:2412.15115}, publisher={arXiv}, author={Qwen and Yang, An and Yang, Baosong and Zhang, Beichen and Hui, Binyuan and Zheng, Bo and Yu, Bowen and Li, Chengyuan and Liu, Dayiheng and Huang, Fei and Wei, Haoran and Lin, Huan and Yang, Jian and Tu, Jianhong and Zhang, Jianwei and Yang, Jianxin and Yang, Jiaxi and Zhou, Jingren and Lin, Junyang and Dang, Kai and Lu, Keming and Bao, Keqin and Yang, Kexin and Yu, Le and Li, Mei and Xue, Mingfeng and Zhang, Pei and Zhu, Qin and Men, Rui and Lin, Runji and Li, Tianhao and Tang, Tianyi and Xia, Tingyu and Ren, Xingzhang and Ren, Xuancheng and Fan, Yang and Su, Yang and Zhang, Yichang and Wan, Yu and Liu, Yuqiong and Cui, Zeyu and Zhang, Zhenru and Qiu, Zihan}, year={2025}, month=jan }

@article{Grattafiori_Dubey_Jauhri_Pandey_Kadian_Al-Dahle_Letman_Mathur_Schelten_Vaughan_et, title={The Llama 3 Herd of Models}, url={http://arxiv.org/abs/2407.21783}, DOI={10.48550/arXiv.2407.21783}, abstractNote={Modern artificial intelligence (AI) systems are powered by foundation models. This paper presents a new set of foundation models, called Llama 3. It is a herd of language models that natively support multilinguality, coding, reasoning, and tool usage. Our largest model is a dense Transformer with 405B parameters and a context window of up to 128K tokens. This paper presents an extensive empirical evaluation of Llama 3. We find that Llama 3 delivers comparable quality to leading language models such as GPT-4 on a plethora of tasks. We publicly release Llama 3, including pre-trained and post-trained versions of the 405B parameter language model and our Llama Guard 3 model for input and output safety. The paper also presents the results of experiments in which we integrate image, video, and speech capabilities into Llama 3 via a compositional approach. We observe this approach performs competitively with the state-of-the-art on image, video, and speech recognition tasks. The resulting models are not yet being broadly released as they are still under development.}, note={arXiv:2407.21783 [cs]}, number={arXiv:2407.21783}, publisher={arXiv}, author={Grattafiori, Aaron and Dubey, Abhimanyu and Jauhri, Abhinav and Pandey, Abhinav and Kadian, Abhishek and Al-Dahle, Ahmad and Letman, Aiesha and Mathur, Akhil and Schelten, Alan and Vaughan, Alex and Yang, Amy and Fan, Angela and Goyal, Anirudh and Hartshorn, Anthony and Yang, Aobo and Mitra, Archi and Sravankumar, Archie and Korenev, Artem and Hinsvark, Arthur and Rao, Arun and Zhang, Aston and Rodriguez, Aurelien and Gregerson, Austen and Spataru, Ava and Roziere, Baptiste and Biron, Bethany and Tang, Binh and Chern, Bobbie and Caucheteux, Charlotte and Nayak, Chaya and Bi, Chloe and Marra, Chris and McConnell, Chris and Keller, Christian and Touret, Christophe and Wu, Chunyang and Wong, Corinne and Ferrer, Cristian Canton and Nikolaidis, Cyrus and Allonsius, Damien and Song, Daniel and Pintz, Danielle and Livshits, Danny and Wyatt, Danny and Esiobu, David and Choudhary, Dhruv and Mahajan, Dhruv and Garcia-Olano, Diego and Perino, Diego and Hupkes, Dieuwke and Lakomkin, Egor and AlBadawy, Ehab and Lobanova, Elina and Dinan, Emily and Smith, Eric Michael and Radenovic, Filip and Guzmán, Francisco and Zhang, Frank and Synnaeve, Gabriel and Lee, Gabrielle and Anderson, Georgia Lewis and Thattai, Govind and Nail, Graeme and Mialon, Gregoire and Pang, Guan and Cucurell, Guillem and Nguyen, Hailey and Korevaar, Hannah and Xu, Hu and Touvron, Hugo and Zarov, Iliyan and Ibarra, Imanol Arrieta and Kloumann, Isabel and Misra, Ishan and Evtimov, Ivan and Zhang, Jack and Copet, Jade and Lee, Jaewon and Geffert, Jan and Vranes, Jana and Park, Jason and Mahadeokar, Jay and Shah, Jeet and Linde, Jelmer van der and Billock, Jennifer and Hong, Jenny and Lee, Jenya and Fu, Jeremy and Chi, Jianfeng and Huang, Jianyu and Liu, Jiawen and Wang, Jie and Yu, Jiecao and Bitton, Joanna and Spisak, Joe and Park, Jongsoo and Rocca, Joseph and Johnstun, Joshua and Saxe, Joshua and Jia, Junteng and Alwala, Kalyan Vasuden and Prasad, Karthik and Upasani, Kartikeya and Plawiak, Kate and Li, Ke and Heafield, Kenneth and Stone, Kevin and El-Arini, Khalid and Iyer, Krithika and Malik, Kshitiz and Chiu, Kuenley and Bhalla, Kunal and Lakhotia, Kushal and Rantala-Yeary, Lauren and Maaten, Laurens van der and Chen, Lawrence and Tan, Liang and Jenkins, Liz and Martin, Louis and Madaan, Lovish and Malo, Lubo and Blecher, Lukas and Landzaat, Lukas and Oliveira, Luke de and Muzzi, Madeline and Pasupuleti, Mahesh and Singh, Mannat and Paluri, Manohar and Kardas, Marcin and Tsimpoukelli, Maria and Oldham, Mathew and Rita, Mathieu and Pavlova, Maya and Kambadur, Melanie and Lewis, Mike and Si, Min and Singh, Mitesh Kumar and Hassan, Mona and Goyal, Naman and Torabi, Narjes and Bashlykov, Nikolay and Bogoychev, Nikolay and Chatterji, Niladri and Zhang, Ning and Duchenne, Olivier and Çelebi, Onur and Alrassy, Patrick and Zhang, Pengchuan and Li, Pengwei and Vasic, Petar and Weng, Peter and Bhargava, Prajjwal and Dubal, Pratik and Krishnan, Praveen and Koura, Punit Singh and Xu, Puxin and He, Qing and Dong, Qingxiao and Srinivasan, Ragavan and Ganapathy, Raj and Calderer, Ramon and Cabral, Ricardo Silveira and Stojnic, Robert and Raileanu, Roberta and Maheswari, Rohan and Girdhar, Rohit and Patel, Rohit and Sauvestre, Romain and Polidoro, Ronnie and Sumbaly, Roshan and Taylor, Ross and Silva, Ruan and Hou, Rui and Wang, Rui and Hosseini, Saghar and Chennabasappa, Sahana and Singh, Sanjay and Bell, Sean and Kim, Seohyun Sonia and Edunov, Sergey and Nie, Shaoliang and Narang, Sharan and Raparthy, Sharath and Shen, Sheng and Wan, Shengye and Bhosale, Shruti and Zhang, Shun and Vandenhende, Simon and Batra, Soumya and Whitman, Spencer and Sootla, Sten and Collot, Stephane and Gururangan, Suchin and Borodinsky, Sydney and Herman, Tamar and Fowler, Tara and Sheasha, Tarek and Georgiou, Thomas and Scialom, Thomas and Speckbacher, Tobias and Mihaylov, Todor and Xiao, Tong and Karn, Ujjwal and Goswami, Vedanuj and Gupta, Vibhor and Ramanathan, Vignesh and Kerkez, Viktor and Gonguet, Vincent and Do, Virginie and Vogeti, Vish and Albiero, Vítor and Petrovic, Vladan and Chu, Weiwei and Xiong, Wenhan and Fu, Wenyin and Meers, Whitney and Martinet, Xavier and Wang, Xiaodong and Wang, Xiaofang and Tan, Xiaoqing Ellen and Xia, Xide and Xie, Xinfeng and Jia, Xuchao and Wang, Xuewei and Goldschlag, Yaelle and Gaur, Yashesh and Babaei, Yasmine and Wen, Yi and Song, Yiwen and Zhang, Yuchen and Li, Yue and Mao, Yuning and Coudert, Zacharie Delpierre and Yan, Zheng and Chen, Zhengxing and Papakipos, Zoe and Singh, Aaditya and Srivastava, Aayushi and Jain, Abha and Kelsey, Adam and Shajnfeld, Adam and Gangidi, Adithya and Victoria, Adolfo and Goldstand, Ahuva and Menon, Ajay and Sharma, Ajay and Boesenberg, Alex and Baevski, Alexei and Feinstein, Allie and Kallet, Amanda and Sangani, Amit and Teo, Amos and Yunus, Anam and Lupu, Andrei and Alvarado, Andres and Caples, Andrew and Gu, Andrew and Ho, Andrew and Poulton, Andrew and Ryan, Andrew and Ramchandani, Ankit and Dong, Annie and Franco, Annie and Goyal, Anuj and Saraf, Aparajita and Chowdhury, Arkabandhu and Gabriel, Ashley and Bharambe, Ashwin and Eisenman, Assaf and Yazdan, Azadeh and James, Beau and Maurer, Ben and Leonhardi, Benjamin and Huang, Bernie and Loyd, Beth and Paola, Beto De and Paranjape, Bhargavi and Liu, Bing and Wu, Bo and Ni, Boyu and Hancock, Braden and Wasti, Bram and Spence, Brandon and Stojkovic, Brani and Gamido, Brian and Montalvo, Britt and Parker, Carl and Burton, Carly and Mejia, Catalina and Liu, Ce and Wang, Changhan and Kim, Changkyu and Zhou, Chao and Hu, Chester and Chu, Ching-Hsiang and Cai, Chris and Tindal, Chris and Feichtenhofer, Christoph and Gao, Cynthia and Civin, Damon and Beaty, Dana and Kreymer, Daniel and Li, Daniel and Adkins, David and Xu, David and Testuggine, Davide and David, Delia and Parikh, Devi and Liskovich, Diana and Foss, Didem and Wang, Dingkang and Le, Duc and Holland, Dustin and Dowling, Edward and Jamil, Eissa and Montgomery, Elaine and Presani, Eleonora and Hahn, Emily and Wood, Emily and Le, Eric-Tuan and Brinkman, Erik and Arcaute, Esteban and Dunbar, Evan and Smothers, Evan and Sun, Fei and Kreuk, Felix and Tian, Feng and Kokkinos, Filippos and Ozgenel, Firat and Caggioni, Francesco and Kanayet, Frank and Seide, Frank and Florez, Gabriela Medina and Schwarz, Gabriella and Badeer, Gada and Swee, Georgia and Halpern, Gil and Herman, Grant and Sizov, Grigory and Guangyi and Zhang and Lakshminarayanan, Guna and Inan, Hakan and Shojanazeri, Hamid and Zou, Han and Wang, Hannah and Zha, Hanwen and Habeeb, Haroun and Rudolph, Harrison and Suk, Helen and Aspegren, Henry and Goldman, Hunter and Zhan, Hongyuan and Damlaj, Ibrahim and Molybog, Igor and Tufanov, Igor and Leontiadis, Ilias and Veliche, Irina-Elena and Gat, Itai and Weissman, Jake and Geboski, James and Kohli, James and Lam, Janice and Asher, Japhet and Gaya, Jean-Baptiste and Marcus, Jeff and Tang, Jeff and Chan, Jennifer and Zhen, Jenny and Reizenstein, Jeremy and Teboul, Jeremy and Zhong, Jessica and Jin, Jian and Yang, Jingyi and Cummings, Joe and Carvill, Jon and Shepard, Jon and McPhie, Jonathan and Torres, Jonathan and Ginsburg, Josh and Wang, Junjie and Wu, Kai and U, Kam Hou and Saxena, Karan and Khandelwal, Kartikay and Zand, Katayoun and Matosich, Kathy and Veeraraghavan, Kaushik and Michelena, Kelly and Li, Keqian and Jagadeesh, Kiran and Huang, Kun and Chawla, Kunal and Huang, Kyle and Chen, Lailin and Garg, Lakshya and A, Lavender and Silva, Leandro and Bell, Lee and Zhang, Lei and Guo, Liangpeng and Yu, Licheng and Moshkovich, Liron and Wehrstedt, Luca and Khabsa, Madian and Avalani, Manav and Bhatt, Manish and Mankus, Martynas and Hasson, Matan and Lennie, Matthew and Reso, Matthias and Groshev, Maxim and Naumov, Maxim and Lathi, Maya and Keneally, Meghan and Liu, Miao and Seltzer, Michael L. and Valko, Michal and Restrepo, Michelle and Patel, Mihir and Vyatskov, Mik and Samvelyan, Mikayel and Clark, Mike and Macey, Mike and Wang, Mike and Hermoso, Miquel Jubert and Metanat, Mo and Rastegari, Mohammad and Bansal, Munish and Santhanam, Nandhini and Parks, Natascha and White, Natasha and Bawa, Navyata and Singhal, Nayan and Egebo, Nick and Usunier, Nicolas and Mehta, Nikhil and Laptev, Nikolay Pavlovich and Dong, Ning and Cheng, Norman and Chernoguz, Oleg and Hart, Olivia and Salpekar, Omkar and Kalinli, Ozlem and Kent, Parkin and Parekh, Parth and Saab, Paul and Balaji, Pavan and Rittner, Pedro and Bontrager, Philip and Roux, Pierre and Dollar, Piotr and Zvyagina, Polina and Ratanchandani, Prashant and Yuvraj, Pritish and Liang, Qian and Alao, Rachad and Rodriguez, Rachel and Ayub, Rafi and Murthy, Raghotham and Nayani, Raghu and Mitra, Rahul and Parthasarathy, Rangaprabhu and Li, Raymond and Hogan, Rebekkah and Battey, Robin and Wang, Rocky and Howes, Russ and Rinott, Ruty and Mehta, Sachin and Siby, Sachin and Bondu, Sai Jayesh and Datta, Samyak and Chugh, Sara and Hunt, Sara and Dhillon, Sargun and Sidorov, Sasha and Pan, Satadru and Mahajan, Saurabh and Verma, Saurabh and Yamamoto, Seiji and Ramaswamy, Sharadh and Lindsay, Shaun and Lindsay, Shaun and Feng, Sheng and Lin, Shenghao and Zha, Shengxin Cindy and Patil, Shishir and Shankar, Shiva and Zhang, Shuqiang and Zhang, Shuqiang and Wang, Sinong and Agarwal, Sneha and Sajuyigbe, Soji and Chintala, Soumith and Max, Stephanie and Chen, Stephen and Kehoe, Steve and Satterfield, Steve and Govindaprasad, Sudarshan and Gupta, Sumit and Deng, Summer and Cho, Sungmin and Virk, Sunny and Subramanian, Suraj and Choudhury, Sy and Goldman, Sydney and Remez, Tal and Glaser, Tamar and Best, Tamara and Koehler, Thilo and Robinson, Thomas and Li, Tianhe and Zhang, Tianjun and Matthews, Tim and Chou, Timothy and Shaked, Tzook and Vontimitta, Varun and Ajayi, Victoria and Montanez, Victoria and Mohan, Vijai and Kumar, Vinay Satish and Mangla, Vishal and Ionescu, Vlad and Poenaru, Vlad and Mihailescu, Vlad Tiberiu and Ivanov, Vladimir and Li, Wei and Wang, Wenchen and Jiang, Wenwen and Bouaziz, Wes and Constable, Will and Tang, Xiaocheng and Wu, Xiaojian and Wang, Xiaolan and Wu, Xilun and Gao, Xinbo and Kleinman, Yaniv and Chen, Yanjun and Hu, Ye and Jia, Ye and Qi, Ye and Li, Yenda and Zhang, Yilin and Zhang, Ying and Adi, Yossi and Nam, Youngjin and Yu and Wang and Zhao, Yu and Hao, Yuchen and Qian, Yundi and Li, Yunlu and He, Yuzi and Rait, Zach and DeVito, Zachary and Rosnbrick, Zef and Wen, Zhaoduo and Yang, Zhenyu and Zhao, Zhiwei and Ma, Zhiyu}, year={2024}, month=nov }

@inproceedings{Wei_Wang_Schuurmans_Bosma_Ichter_Xia_Chi_Le_Zhou_2022, address={Red Hook, NY, USA}, series={NIPS ’22}, title={Chain-of-thought prompting elicits reasoning in large language models}, ISBN={978-1-7138-7108-8}, abstractNote={We explore how generating a chain of thought—a series of intermediate reasoning steps—significantly improves the ability of large language models to perform complex reasoning. In particular, we show how such reasoning abilities emerge naturally in sufficiently large language models via a simple method called chain-of-thought prompting, where a few chain of thought demonstrations are provided as exemplars in prompting.Experiments on three large language models show that chain-of-thought prompting improves performance on a range of arithmetic, commonsense, and symbolic reasoning tasks. The empirical gains can be striking. For instance, prompting a PaLM 540B with just eight chain-of-thought exemplars achieves state-of-the-art accuracy on the GSM8K benchmark of math word problems, surpassing even finetuned GPT-3 with a verifier.}, booktitle={Proceedings of the 36th International Conference on Neural Information Processing Systems}, publisher={Curran Associates Inc.}, author={Wei, Jason and Wang, Xuezhi and Schuurmans, Dale and Bosma, Maarten and Ichter, Brian and Xia, Fei and Chi, Ed H. and Le, Quoc V. and Zhou, Denny}, year={2022}, month=nov, pages={24824–24837}, collection={NIPS ’22} }

@inproceedings{Kojima_Gu_Reid_Matsuo_Iwasawa_2022, address={Red Hook, NY, USA}, series={NIPS ’22}, title={Large language models are zero-shot reasoners}, ISBN={978-1-7138-7108-8}, abstractNote={Pretrained large language models (LLMs) are widely used in many sub-fields of natural language processing (NLP) and generally known as excellent few-shot learners with task-specific exemplars. Notably, chain of thought (CoT) prompting, a recent technique for eliciting complex multi-step reasoning through step-by-step answer examples, achieved the state-of-the-art performances in arithmetics and symbolic reasoning, difficult system-2 tasks that do not follow the standard scaling laws for LLMs. While these successes are often attributed to LLMs’ ability for few-shot learning, we show that LLMs are decent zero-shot reasoners by simply adding “Let’s think step by step” before each answer. Experimental results demonstrate that our Zero-shot-CoT, using the same single prompt template, significantly outperforms zero-shot LLM performances on diverse benchmark reasoning tasks including arithmetics (MultiArith, GSM8K, AQUA-RAT, SVAMP), symbolic reasoning (Last Letter, Coin Flip), and other logical reasoning tasks (Date Understanding, Tracking Shuffled Objects), without any hand-crafted few-shot examples, e.g. increasing the accuracy on MultiArith from 17.7% to 78.7% and GSM8K from 10.4% to 40.7% with large-scale InstructGPT model (text-davinci-002), as well as similar magnitudes of improvements with another off-the-shelf large model, 540B parameter PaLM. The versatility of this single prompt across very diverse reasoning tasks hints at untapped and understudied fundamental zero-shot capabilities of LLMs, suggesting high-level, multi-task broad cognitive capabilities may be extracted by simple prompting. We hope our work not only serves as the minimal strongest zero-shot baseline for the challenging reasoning benchmarks, but also highlights the importance of carefully exploring and analyzing the enormous zero-shot knowledge hidden inside LLMs before crafting finetuning datasets or few-shot exemplars.}, booktitle={Proceedings of the 36th International Conference on Neural Information Processing Systems}, publisher={Curran Associates Inc.}, author={Kojima, Takeshi and Gu, Shixiang Shane and Reid, Machel and Matsuo, Yutaka and Iwasawa, Yusuke}, year={2022}, month=nov, pages={22199–22213}, collection={NIPS ’22}, url={https://dl.acm.org/doi/10.5555/3600270.3601883} }

@misc{zeff2024gpt4omini,
  author       = {Zeff, Maxwell},
  title        = {{OpenAI} Unveils {GPT-4o} Mini, a Smaller and Cheaper {AI} Model},
  howpublished = {TechCrunch},
  year         = {2024},
  month        = jul,
  day          = {18},
  url          = {https://techcrunch.com/2024/07/18/openai-unveils-gpt-4o-mini-a-small-ai-model-powering-chatgpt/},
  note         = {Accessed: 2026-02-20}
}

@inproceedings{Brown_Mann_Ryder_Subbiah_Kaplan_Dhariwal_Neelakantan_Shyam_Sastry_Askell_2020, title={Language Models are Few-Shot Learners}, volume={33}, url={https://papers.nips.cc/paper/2020/hash/1457c0d6bfcb4967418bfb8ac142f64a-Abstract.html}, abstractNote={We demonstrate that scaling up language models greatly improves task-agnostic, few-shot performance, sometimes even becoming competitive with prior state-of-the-art fine-tuning approaches. Specifically, we train GPT-3, an autoregressive language model with 175 billion parameters, 10x more than any previous non-sparse language model, and test its performance in the few-shot setting.  For all tasks, GPT-3 is applied without any gradient updates or fine-tuning, with tasks and few-shot demonstrations specified purely via text interaction with the model.  GPT-3 achieves strong performance on many NLP datasets, including translation, question-answering, and cloze tasks. We also identify some datasets where GPT-3’s few-shot learning still struggles, as well as some datasets where GPT-3 faces methodological issues related to training on large web corpora.}, booktitle={Advances in Neural Information Processing Systems}, publisher={Curran Associates, Inc.}, author={Brown, Tom and Mann, Benjamin and Ryder, Nick and Subbiah, Melanie and Kaplan, Jared D and Dhariwal, Prafulla and Neelakantan, Arvind and Shyam, Pranav and Sastry, Girish and Askell, Amanda and Agarwal, Sandhini and Herbert-Voss, Ariel and Krueger, Gretchen and Henighan, Tom and Child, Rewon and Ramesh, Aditya and Ziegler, Daniel and Wu, Jeffrey and Winter, Clemens and Hesse, Chris and Chen, Mark and Sigler, Eric and Litwin, Mateusz and Gray, Scott and Chess, Benjamin and Clark, Jack and Berner, Christopher and McCandlish, Sam and Radford, Alec and Sutskever, Ilya and Amodei, Dario}, year={2020}, pages={1877–1901} }

@InProceedings{pmlr-v239-mohta23a,
  title = 	 {Are large language models good annotators?},
  author =       {Mohta, Jay and Ak, Kenan and Xu, Yan and Shen, Mingwei},
  booktitle = 	 {Proceedings on "I Can't Believe It's Not Better: Failure  Modes in the Age of Foundation Models" at NeurIPS 2023 Workshops},
  pages = 	 {38--48},
  year = 	 {2023},
  editor = 	 {Antorán, Javier and Blaas, Arno and Buchanan, Kelly and Feng, Fan and Fortuin, Vincent and Ghalebikesabi, Sahra and Kriegler, Andreas and Mason, Ian and Rohde, David and Ruiz, Francisco J. R. and Uelwer, Tobias and Xie, Yubin and Yang, Rui},
  volume = 	 {239},
  series = 	 {Proceedings of Machine Learning Research},
  month = 	 {16 Dec},
  publisher =    {PMLR},
  url = 	 {https://proceedings.mlr.press/v239/mohta23a.html}
}

@softwareversion{plique2019minet,
  TITLE = {{Minet, a webmining CLI tool \& library for python.}},
  AUTHOR = {Plique, Guillaume and Breteau, Pauline and Farjas, Jules and Th{\'e}ro, H{\'e}lo{\"i}se and Descamps, Jean and Pell{\'e}, Am{\'e}lie and Miguel, Laura},
  URL = {https://sciencespo.hal.science/hal-03903317},
  NOTE = {},
  YEAR = {2019},
  MONTH = Feb,
  DOI = {10.5281/zenodo.4564399},
  SWHID = {swh:1:dir:f8668a6334770363869152ed70214d8b21771fd5;origin=https://hal.archives-ouvertes.fr/hal-03903317;visit=swh:1:snp:5e5713fe007b2fe71c4f04ca3211ff33c6e5cb2e;anchor=swh:1:rel:c5bde112d067ecae4cffbcd3576a6a47f7f37210;path=/},
  VERSION = {0.66.1},
  REPOSITORY = {https://github.com/medialab/minet},
  LICENSE = {GNU General Public License v3.0 only},
  HAL_ID = {hal-03903317},
  HAL_VERSION = {v1},
}

@article{shu2017fake, title={Fake News Detection on Social Media: A Data Mining Perspective}, volume={19}, ISSN={1931-0145}, DOI={10.1145/3137597.3137600}, abstractNote={Social media for news consumption is a double-edged sword. On the one hand, its low cost, easy access, and rapid dissemination of information lead people to seek out and consume news from social media. On the other hand, it enables the wide spread of fake news", i.e., low quality news with intentionally false information. The extensive spread of fake news has the potential for extremely negative impacts on individuals and society. Therefore, fake news detection on social media has recently become an emerging research that is attracting tremendous attention. Fake news detection on social media presents unique characteristics and challenges that make existing detection algorithms from traditional news media ine ective or not applicable. First, fake news is intentionally written to mislead readers to believe false information, which makes it difficult and nontrivial to detect based on news content; therefore, we need to include auxiliary information, such as user social engagements on social media, to help make a determination. Second, exploiting this auxiliary information is challenging in and of itself as users’ social engagements with fake news produce data that is big, incomplete, unstructured, and noisy. Because the issue of fake news detection on social media is both challenging and relevant, we conducted this survey to further facilitate research on the problem. In this survey, we present a comprehensive review of detecting fake news on social media, including fake news characterizations on psychology and social theories, existing algorithms from a data mining perspective, evaluation metrics and representative datasets. We also discuss related research areas, open problems, and future research directions for fake news detection on social media.}, number={1}, journal={SIGKDD Explor. Newsl.}, author={Shu, Kai and Sliva, Amy and Wang, Suhang and Tang, Jiliang and Liu, Huan}, year={2017}, month=sept, pages={22–36} }

@inproceedings{bender2011annotating, address={Portland, Oregon}, title={Annotating Social Acts: Authority Claims and Alignment Moves in Wikipedia Talk Pages}, url={https://aclanthology.org/W11-0707/}, booktitle={Proceedings of the Workshop on Language in Social Media (LSM 2011)}, publisher={Association for Computational Linguistics}, author={Bender, Emily M. and Morgan, Jonathan T. and Oxley, Meghan and Zachry, Mark and Hutchinson, Brian and Marin, Alex and Zhang, Bin and Ostendorf, Mari}, editor={Nagarajan, Meenakshi and Gamon, Michael}, year={2011}, month=june, pages={48–57} }

@inproceedings{walker2012corpus, address={Istanbul, Turkey}, title={A Corpus for Research on Deliberation and Debate}, url={https://aclanthology.org/L12-1643/}, abstractNote={Deliberative, argumentative discourse is an important component of opinion formation, belief revision, and knowledge discovery; it is a cornerstone of modern civil society. Argumentation is productively studied in branches ranging from theoretical artificial intelligence to political rhetoric, but empirical analysis has suffered from a lack of freely available, unscripted argumentative dialogs. This paper presents the Internet Argument Corpus (IAC), a set of 390,704 posts in 11,800 discussions extracted from the online debate site 4forums.com. A 2866 thread/130,206 post extract of the corpus has been manually sided for topic of discussion, and subsets of this topic-labeled extract have been annotated for several dialogic and argumentative markers: degrees of agreement with a previous post, cordiality, audience-direction, combativeness, assertiveness, emotionality of argumentation, and sarcasm. As an application of this resource, the paper closes with a discussion of the relationship between discourse marker pragmatics, agreement, emotionality, and sarcasm in the IAC corpus.}, booktitle={Proceedings of the Eighth International Conference on Language Resources and Evaluation (LREC’12)}, publisher={European Language Resources Association (ELRA)}, author={Walker, Marilyn and Tree, Jean Fox and Anand, Pranav and Abbott, Rob and King, Joseph}, editor={Calzolari, Nicoletta and Choukri, Khalid and Declerck, Thierry and Doğan, Mehmet Uğur and Maegaard, Bente and Mariani, Joseph and Moreno, Asuncion and Odijk, Jan and Piperidis, Stelios}, year={2012}, month=may, pages={812–817} }

@article{zhang2017characterizing, title={Characterizing Online Discussion Using Coarse Discourse Sequences}, volume={11}, rights={Copyright (c) 2021 Proceedings of the International AAAI Conference on Web and Social Media}, ISSN={2334-0770}, DOI={10.1609/icwsm.v11i1.14886}, abstractNote={In this work, we present a novel method for classifying comments in online discussions into a set of coarse discourse acts towards the goal of better understanding discussions at scale. To facilitate this study, we devise a categorization of coarse discourse acts designed to encompass general online discussion and allow for easy annotation by crowd workers. We collect and release a corpus of over 9,000 threads comprising over 100,000 comments manually annotated via paid crowdsourcing with discourse acts and randomly sampled from the site Reddit. Using our corpus, we demonstrate how the analysis of discourse acts can characterize different types of discussions, including discourse sequences such as Q&amp;A pairs and chains of disagreement, as well as different communities. Finally, we conduct experiments to predict discourse acts using our corpus, finding that structured prediction models such as conditional random fields can achieve an F1 score of 75%. We also demonstrate how the broadening of discourse acts from simply question and answer to a richer set of categories can improve the recall performance of Q&amp;A extraction.}, number={1}, journal={Proceedings of the International AAAI Conference on Web and Social Media}, author={Zhang, Amy and Culbertson, Bryan and Paritosh, Praveen}, year={2017}, month=may, pages={357–366}, language={en} }

@inproceedings{pougue2021debagreement,
 author = {Pougu\'{e}-Biyong, John and Semenova, Valentina and Matton, Alexandre and Han, Rachel and Kim, Aerin and Lambiotte, Renaud and Farmer, Doyne},
 booktitle = {Proceedings of the Neural Information Processing Systems Track on Datasets and Benchmarks},
 editor = {J. Vanschoren and S. Yeung},
 pages = {},
 title = {DEBAGREEMENT: A comment-reply dataset for (dis)agreement detection in online debates},
 url = {https://datasets-benchmarks-proceedings.neurips.cc/paper_files/paper/2021/file/6f3ef77ac0e3619e98159e9b6febf557-Paper-round2.pdf},
 volume = {1},
 year = {2021}
}

@inproceedings{poudhar2024strategy, address={Mexico City, Mexico}, title={A Strategy Labelled Dataset of Counterspeech}, url={https://aclanthology.org/2024.woah-1.20/}, DOI={10.18653/v1/2024.woah-1.20}, abstractNote={Increasing hateful conduct online demands effective counterspeech strategies to mitigate its impact. We introduce a novel dataset annotated with such strategies, aimed at facilitating the generation of targeted responses to hateful language. We labelled 1000 hate speech/counterspeech pairs from an existing dataset with strategies established in the social sciences. We find that a one-shot prompted classification model achieves promising accuracy in classifying the strategies according to the manual labels, demonstrating the potential of generative Large Language Models (LLMs) to distinguish between counterspeech strategies.}, booktitle={Proceedings of the 8th Workshop on Online Abuse and Harms (WOAH 2024)}, publisher={Association for Computational Linguistics}, author={Poudhar, Aashima and Konstas, Ioannis and Abercrombie, Gavin}, editor={Chung, Yi-Ling and Talat, Zeerak and Nozza, Debora and Plaza-del-Arco, Flor Miriam and Röttger, Paul and Mostafazadeh Davani, Aida and Calabrese, Agostina}, year={2024}, month=june, pages={256–265} }

@article{zubiaga2016pheme,
    author = "Arkaitz Zubiaga and Geraldine Wong Sak Hoi and Maria Liakata and Rob Procter",
    title = "{PHEME dataset of rumours and non-rumours}",
    year = "2016",
    month = "10",
    url = "https://figshare.com/articles/dataset/PHEME_dataset_of_rumours_and_non-rumours/4010619",
    journal={arXiv preprint},
    doi = "10.6084/m9.figshare.4010619.v1"
}

@inproceedings{derczynski2017semeval, address={Vancouver, Canada}, title={SemEval-2017 Task 8: RumourEval: Determining rumour veracity and support for rumours}, url={https://aclanthology.org/S17-2006/}, DOI={10.18653/v1/S17-2006}, abstractNote={Media is full of false claims. Even Oxford Dictionaries named “post-truth” as the word of 2016. This makes it more important than ever to build systems that can identify the veracity of a story, and the nature of the discourse around it. RumourEval is a SemEval shared task that aims to identify and handle rumours and reactions to them, in text. We present an annotation scheme, a large dataset covering multiple topics – each having their own families of claims and replies – and use these to pose two concrete challenges as well as the results achieved by participants on these challenges.}, booktitle={Proceedings of the 11th International Workshop on Semantic Evaluation (SemEval-2017)}, publisher={Association for Computational Linguistics}, author={Derczynski, Leon and Bontcheva, Kalina and Liakata, Maria and Procter, Rob and Wong Sak Hoi, Geraldine and Zubiaga, Arkaitz}, editor={Bethard, Steven and Carpuat, Marine and Apidianaki, Marianna and Mohammad, Saif M. and Cer, Daniel and Jurgens, David}, year={2017}, month=aug, pages={69–76} }

@article{fleiss1971measuring,
  title={Measuring nominal scale agreement among many raters},
  author={Fleiss, Joseph L.},
  journal={Psychological Bulletin},
  volume={76},
  number={5},
  pages={378--382},
  year={1971},
  doi={10.1037/h0031619}
}

@book{Lull_1995,
  author    = {Lull, James},
  title     = {Media, Communication, Culture: A Global Approach},
  publisher = {Columbia University Press},
  year      = {1995}
}

\newpage
\clearpage
\appendix
\renewcommand{\thesection}{\Alph{section}} 

\section{Annotation Details}
\label{appendix:annotation}

\paragraph{Annotator Demographics}
The main annotator is a female researcher in Sociology from France, aged 25--35. Two additional annotators participated in calibration: (1) a male researcher in Sociology from France, aged 35--45, and (2) a female researcher in NLP and Computational Social Science from France, aged 25--35.

\paragraph{Pilot Testing and Guideline Calibration}
To assess clarity and applicability of the annotation guidelines, a pilot study was conducted on deliberately challenging sets of ten comments selected by the main annotator. Annotators independently labeled each item and provided brief justification notes. Disagreements and the underlying rationales were discussed during group calibration sessions. This iterative process led to targeted refinements of the guidelines, including explicit instructions for handling irony and parody, rules for interpreting very short text fragments in context, and a conservative default rule when contextual evidence was insufficient. The process was repeated twice, until annotators reached a high level of consistency in applying the rules. Detailed information about the calibration sets are provided in the \href{https://github.com/celia-nouri/SPOT-benchmark/blob/main/Annotation_Guidelines.pdf}{Annotation Guidelines}.

\section{Context Concatenation}  
\label{appendix:context}
To integrate contextual information, all available elements of the publication for each comment were concatenated into a single input string. Each element was prepended with a descriptive tag indicating its type: \texttt{[PARENT]} for the parent comment, \texttt{[POST]} for the post message, \texttt{[ARTICLE]} for the article title and description, \texttt{[ACCOUNT]} for the page or group name, and \texttt{[DOMAIN]} for the media source. Elements were joined using the model's \texttt{[SEP]} token.  

Before concatenation, each field was truncated according to empirically defined limits based on its typical length: comment text to 600 tokens, post title to 200 tokens, article title and description to 200 tokens each, parent comment to 300 tokens, account name to 50 tokens, and domain name to 50 tokens. Empty fields were omitted from the concatenated string to reduce noise. The final sequence was truncated to fit within \textsc{CamemBERT}'s 512-token input window, allowing the model to leverage rich contextual cues while respecting input size constraints. We use the same truncation rules when providing contextual elements to the LLMs prompts to ensure comparability. 

\section{Prompts Used for LLM Experiments}
\label{appendix:prompts}

This section presents the prompts employed in our LLM experiments.

\subsection{Prompt 1: Stopping Point Detection Based on the Annotation Guidelines}

\begin{lstlisting}
Rôle et tâche :
Vous êtes un annotateur. Votre tâche est de déterminer si un commentaire Facebook est un point d’arrêt (Oui) ou non (Non).
Chaque commentaire est associé à un post contenant un lien ou un article signalé pour fausse information. Le commentaire peut être directement sous le post ou être une réponse à un autre commentaire. Pour cette tâche, vous n’avez pas accès à ce contexte de publication.

Définition :
Un point d’arrêt est une intervention critique dans une conversation en ligne.  
Cela peut aller d’une simple expression de doute à une réfutation ou un appel à vérification.  
Un commentaire peut être un point d’arrêt s’il critique, corrige, émet un doute ou demande une vérification sur :
- la crédibilité ou la fiabilité du contenu,
- la pertinence ou la forme (texte, image, vidéo),
- la source ou l’auteur du post,
- les autres utilisateurs.

Cas particuliers :
- Réfutation implicite (sans mots-clés) : Oui si opposition, critique ou avertissement ; Non si simple ajout d’info.
- Alignement incrédule (“c’est peut-être faux mais je m’en fiche”) : Non si émotion/adhésion seulement ; Oui si demande de vérification ou doute explicite.
- Ironie/parodie/sarcasme : Non si purement humoristique ou phatique ; Oui si utilisé pour réfuter, corriger ou exprimer une incrédulité critique.

Instructions :
- Déterminez si le commentaire suivant est un point d’arrêt (Oui) ou non (Non).
- Répondez uniquement par Oui ou Non.

Commentaire : « {text} »
Réponse :
\end{lstlisting}

\subsection{Prompt 2: Annotation Guidelines with Few-Shot Examples}
\begin{lstlisting}
Rôle et tâche :
Vous êtes un annotateur. Votre tâche est de déterminer si un commentaire Facebook est un point d’arrêt (Oui) ou non (Non).
Chaque commentaire est associé à un post contenant un lien ou un article signalé pour fausse information. Le commentaire peut être directement sous le post ou être une réponse à un autre commentaire. Pour cette tâche, vous n’avez pas accès à ce contexte de publication.

Définition :
Un point d’arrêt est une intervention critique dans une conversation en ligne.  
Cela peut aller d’une simple expression de doute à une réfutation ou un appel à vérification.  
Un commentaire peut être un point d’arrêt s’il critique, corrige, émet un doute ou demande une vérification sur :
- la crédibilité ou la fiabilité du contenu,
- la pertinence ou la forme (texte, image, vidéo),
- la source ou l’auteur du post,
- les autres utilisateurs.

Exemples :
« Ah oui, bien sûr on vous croit... » → Oui (ironie exprimant le doute)
« Encore une rumeur Twitter » → Oui (réfutation implicite)
« Si seulement c’était vrai... » → Non (souhait sans critique)
« C’est dégueulasse » → Non (émotion ou indignation sans remise en cause)

Commentaire : « C’est complètement faux ! / fake news »
Réponse : Oui
Commentaire : « Tellement drôle haha ! »
Réponse : Non
Commentaire : « Arrêtez d’inventer des trucs pareils »
Réponse : Oui
Commentaire : « J’y crois pas une seconde »
Réponse : Oui
Commentaire : « Haha les gens sont fous »
Réponse : Non
Commentaire : « T’as vérifié avant de poster ? »
Réponse : Oui
Commentaire : « Grave, c’est choquant ! »
Réponse : Non
Commentaire : « Ce site raconte toujours n’importe quoi »
Réponse : Oui
Commentaire : « Et sinon, il fait beau chez vous ? »
Réponse : Non
Commentaire : « Encore une intox, sérieux... »
Réponse : Oui

Instructions :
- Déterminez si le commentaire suivant est un point d’arrêt (Oui) ou non (Non).
- Répondez uniquement par Oui ou Non.

Commentaire : « {text} »
Réponse :
\end{lstlisting}

\subsection{Prompt 3: Annotation Guidelines with Publication Context}
\begin{lstlisting}
Rôle et tâche :
Vous êtes un annotateur. Votre tâche est de déterminer si un commentaire Facebook est un point d’arrêt (Oui) ou non (Non), en vous appuyant sur le contexte de publication.
Chaque commentaire est associé à un post contenant un lien ou un article signalé pour fausse information. Le commentaire peut être directement sous le post ou être une réponse à un autre commentaire. Pour cette tâche, vous avez accès au post, lien ou article partagé, à la source médiatique, au commentaire parent, et la page ou le compte ayant publié le contenu.

Définition :
Un point d’arrêt est une intervention critique dans une conversation en ligne.  
Cela peut aller d’une simple expression de doute à une réfutation ou un appel à vérification.  
Un commentaire peut être un point d’arrêt s’il critique, corrige, émet un doute ou demande une vérification sur :
- la crédibilité ou la fiabilité du contenu,
- la pertinence ou la forme (texte, image, vidéo),
- la source ou l’auteur du post,
- les autres utilisateurs.

Cas particuliers :
- Réfutation implicite (sans mots-clés) : Oui si opposition, critique ou avertissement ; Non si simple ajout d’info.
- Alignement incrédule (« c’est peut-être faux mais je m’en fiche ») : Non si émotion/adhésion seulement ; Oui si demande de vérification ou doute explicite.
- Ironie/parodie/sarcasme : Non si purement humoristique ou phatique ; Oui si utilisé pour réfuter, corriger ou exprimer une incrédulité critique.

Instructions :
- Déterminez si le commentaire suivant est un point d’arrêt (Oui) ou non (Non), en tenant compte du contexte de publication (post, article, commentaire parent).
- Répondez uniquement par Oui ou Non.

Post : « {account} : {title} »  
Article partagé : « {domain} : {url_title} {description} » 
Commentaire parent : « {parent_comment} »  
Commentaire : « {text} »
Réponse :
\end{lstlisting}

\subsection{Prompt 4: Annotation Guidelines with Publication Context and Few-Shot Examples}

\begin{lstlisting}
Rôle et tâche :
Vous êtes un annotateur. Votre tâche est de déterminer si un commentaire Facebook est un point d’arrêt (Oui) ou non (Non), en vous appuyant sur le contexte de publication.
Chaque commentaire est associé à un post contenant un lien ou un article signalé pour fausse information. Le commentaire peut être directement sous le post ou être une réponse à un autre commentaire. Pour cette tâche, vous avez accès au post, lien ou article partagé, à la source médiatique, au commentaire parent, et la page ou le compte ayant publié le contenu.

Définition :
Un point d’arrêt est une intervention critique dans une conversation en ligne.  
Cela peut aller d’une simple expression de doute à une réfutation ou un appel à vérification.  
Un commentaire peut être un point d’arrêt s’il critique, corrige, émet un doute ou demande une vérification sur :
- la crédibilité ou la fiabilité du contenu,
- la pertinence ou la forme (texte, image, vidéo),
- la source ou l’auteur du post,
- les autres utilisateurs.

Exemples :
« Ah oui, bien sûr on vous croit ... » → Oui (ironie exprimant le doute)
« Encore une rumeur Twitter » → Oui (réfutation implicite)
« Si seulement c’était vrai... » → Non (souhait, pas critique)
« C’est dégueulasse » → Non (outrage ou émotion sans critique)

Post : « Info Vaccins France : une nouvelle victime des industries pharmaceutiques... RIP »
Article partagé : « lesmoutonsrebelles.com : Encore un bébé de deux mois qui décède 48H après avoir reçu 8 vaccins »
Commentaire parent : « Mais sérieux?! Ils comprennent rien de rien! Il en faudra combien de victimes sérieux! »
Commentaire : « Avant les vaccins c était des milliers de victimes et d'enfants qui mouraient.. faudra pas venir pleurer après! »
Réponse : Oui

Post : « LOSC : Le monde du foot est en deuil ! »
Article partagé : « losc.fr : Un grand joueur s’éteint, le monde du football en pleure »
Commentaire parent : «»
Commentaire : « Toutes ces pages de pub pour annoncer (en retard) que Stéphane Paille est décédé ! Le 27 juin.»
Réponse : Oui

Post : « 1 Million de J'aime Contre Emmanuel Macron : »
Article partagé : « valeursactuelles.com : Parlement européen : Bayer-Monsanto finance bien le parti de Macron »
Commentaire parent : « @USER et oui ... »
Commentaire : « On comprend mieux pourquoi nous mangeons encore du cancer »
Réponse : Non

Post : « Force gilet jaune 31 : »
Article partagé : «planetes360.fr : « Je demande des efforts aux Français »... 14 chauffeurs, 60 cuisiniers et hôteliers : les chiffres sur le cabinet d’Édouard Philippe - PLANETES360 »
Commentaire parent : «»
Commentaire : « Honteux... D’émission !!! »
Réponse : Non

Instructions :
- Déterminez si le commentaire suivant est un point d’arrêt (Oui) ou non (Non), en tenant compte du contexte de publication (post, article, commentaire parent).
- Répondez uniquement par Oui ou Non.

Post : « {account} : {title} »  
Article partagé : « {domain} : {url_title} {description} » 
Commentaire parent : « {parent_comment} »  
Commentaire : « {text} »
Réponse : 
\end{lstlisting}

\subsection{Prompt 5: Annotation Guide with Publication Context and Chain-of-Thought Reasoning}

\begin{lstlisting}
Rôle et tâche :
Vous êtes un annotateur. Votre tâche est de déterminer si un commentaire Facebook est un point d’arrêt (Oui) ou non (Non), en vous appuyant sur le contexte de publication.
Chaque commentaire est associé à un post contenant un lien ou un article signalé pour fausse information. Le commentaire peut être directement sous le post ou être une réponse à un autre commentaire. Pour cette tâche, vous avez accès au post, lien ou article partagé, à la source médiatique, au commentaire parent, et la page ou le compte ayant publié le contenu.

Définition :
Un point d’arrêt est une intervention critique dans une conversation en ligne.  
Cela peut aller d’une simple expression de doute à une réfutation ou un appel à vérification.  
Un commentaire peut être un point d’arrêt s’il critique, corrige, émet un doute ou demande une vérification sur :
- la crédibilité ou la fiabilité du contenu,
- la pertinence ou la forme (texte, image, vidéo),
- la source ou l’auteur du post,
- les autres utilisateurs.

Chaîne de raisonnement :
Étape 1 - Identifier le niveau du commentaire  
- Parent vide → commentaire de premier niveau (Étape 2A).  
- Parent présent → commentaire de second niveau (Étape 2B).  

Étape 2A - Commentaire de premier niveau  
- Si le post et l'article se contredisent :  
  Oui → si le commentaire apporte un élément de critique, de preuve, de doute ou de signalement sur le contenu partagé ou sa lecture.  
  Non → si le commentaire approuve simplement le post (ex. “Exactement !”).  
  - Si le commentaire met en doute, corrige ou critique le contenu, la source ou la page → Oui.  
- Si le commentaire réfute implicitement la position du post (ex. post anti-vax / commentaire pro-vax) → Oui.  
- Si le commentaire mentionne la source pour en questionner la fiabilité → Oui.  
- Si le commentaire se moque du site ou de la page (ex. “Encore une fake news de...”) → Oui.  
- Sinon, ou si le commentaire approuve, exprime une émotion ou est hors sujet → Non.  

Étape 2B — Commentaire de second niveau  
- Si le parent soutient le post → Oui si le commentaire le contredit ou le critique.  
- Si le parent critique le post → Oui si le commentaire l’appuie avec un nouvel argument ; Non s’il se contente d’approuver.  
- Sinon, ou si le commentaire attaque le parent critique → Non (contre-stop).  

Instructions :
- Déterminez si le commentaire suivant est un point d’arrêt (Oui) ou non (Non), en tenant compte du contexte de publication (post, article, commentaire parent).
- Suivez rigoureusement la chaîne de raisonnement ci-dessus avant de répondre.
- Répondez uniquement par Oui ou Non.

Post : « {account} : {title} »  
Article partagé : « {domain} : {url_title} {description} » 
Commentaire parent : « {parent_comment} »  
Commentaire : « {text} »
Réponse :

\end{lstlisting}

\section{Training Setup and Context Concatenation}
\label{appendix:train}

For all encoder-based experiments, the dataset was divided into training, validation, and test splits following an 80/20 train–test ratio, and an additional 80/20 split on the training portion to create the validation set.  
Since stopping points constitute approximately 10\% of the annotated corpus, all splits were stratified to preserve this class distribution.  

Models were trained using a weighted cross-entropy loss to compensate for class imbalance (approximately 10\% positive samples).  
Training was performed for 25 epochs using the AdamW optimizer with a learning rate of $2{\times}10^{-5}$, weight decay of $0.01$, and a batch size of 4 with gradient accumulation over 8 steps to yield an effective batch size of 32.  
All experiments were conducted on distributed multi-GPU setups (NVIDIA RTX 8000 and H100), with fixed random seeds to ensure full reproducibility.

\section{LLM results}
\label{appendix:llm-results}

\begin{table}[!ht]
\begin{center}
\begin{tabular}{|l|c|c|c|c|c|c|c|}
\hline
\textbf{Model} & \textbf{z|f|c} & \textbf{ctx?} & \textbf{$F_{1}$}  & \textbf{Prec.} & \textbf{Recall} & \textbf{Inval} \\
\hline
\textsc{Qwen} & z & n & 39.2 & 67.3 & 27.7 & 0 \\
\hline
\textsc{Qwen} & f & n & 53.6 & 54.2 & 53.1 & 2 \\
\hline
\textsc{Qwen} & z & y & 45.6 & 57.0 & 38.0 & 0 \\
\hline
\textsc{Qwen} & f & y & 42.5 & 34.5 & 55.4 & 6 \\
\hline
\textsc{Qwen} & c & y & 45.6 & 49.1 & 42.5 & 0 \\
\hline
\textsc{Mistral} & z & n & 36.1 & 55.6 & 26.8 & 385 \\
\hline
\textsc{Mistral} & f & n & 44.0 & 33.9 & 62.7 & 1848 \\
\hline
\textsc{Mistral} & f & n & 43.0 & 32.0 & 65.3 & 1637 \\
\hline
\textsc{Mistral} & z & y & 27.8 & 21.8 & 38.2 & 692 \\
\hline
\textsc{Mistral} & f & y & 25.7 & 16.0 & 64.8 & 898 \\
\hline
\textsc{Mistral} & c & y & 31.8 & 23.0 & 51.6 & 963 \\
\hline
\textsc{Llama} & z & n & 24.1 & 15.3 & 56.6 & 0 \\
\hline
\textsc{Llama}  & f & n & 22.4 & 12.8 & 88.5 & 2 \\
\hline
\textsc{Llama} & z & y & 16.4 & 9.0 & 93.6 & 1656 \\
\hline
\textsc{Llama}  & f & y & 20.2 & 11.4 & 90.6 & 9 \\
\hline
\textsc{Llama}  & c & y & 15.1 & 8.5 & 65.3 & 5180 \\
\hline
\textsc{GPT} & z & n & 53.4 & 57.8 & 49.6 & 0 \\ 
\hline
\textsc{GPT} & f & n & \textbf{62.9} & 60.5 & 65.6 & 0 \\ 
\hline
\textsc{GPT} & z & y & 55.9 & 55.6 & 56.2 & 0 \\ 
\hline
\textsc{GPT} & f & y & 55.6 & 45.7 & \textbf{71.0} & 0 \\ 
\hline
\textsc{GPT} & c & y & 54.3 & \textbf{61.3} & 48.8 & 0 \\
\hline
\end{tabular}
\caption{Extended results of LLMs under different prompting strategies ('z' for zero-shot, 'f' for few-shot, and 'c' for chain-of-thought) and contextual settings ('y' for 'with context', and 'n' for 'without context') . Invalid outputs indicate the number of cases where the model failed to produce a valid label.}
\label{app:llm-results}
\end{center}
\end{table}

\end{document}